\documentclass[lettersize,journal]{IEEEtran}
\usepackage{amsmath,amsfonts}
\usepackage{algorithmic}
\usepackage{array}
\usepackage[caption=false,font=normalsize,labelfont=sf,textfont=sf]{subfig}
\usepackage{textcomp}
\usepackage{stfloats}
\usepackage{url}
\usepackage{verbatim}
\usepackage{graphicx}
\def\BibTeX{{\rm B\kern-.05em{\sc i\kern-.025em b}\kern-.08em
    T\kern-.1667em\lower.7ex\hbox{E}\kern-.125emX}}
\usepackage{balance}

\usepackage{cite}

\usepackage{amsfonts}

\usepackage{booktabs}
\usepackage{multirow}
\newcommand{\our}{B-LoRA-XS}
\newcommand{\citep}{\cite}
\newcommand{\tB}{B}
\usepackage{mathabx}
\usepackage{xcolor}

\begin{document}


\title{Bayesian Fine-tuning in Projected Subspaces}

\author{Viktar~Dubovik,
        Patryk~Marsza{\l}ek,
        Jacek~Tabor,
        and~Tomasz~Ku{\'s}mierczyk$^{\dagger}$%
\IEEEcompsocitemizethanks{
    \IEEEcompsocthanksitem V. Dubovik, P. Marsza{\l}ek, J. Tabor, and T. Ku{\'s}mierczyk are with the Jagiellonian University, Kraków, Poland.
    \protect
    \IEEEcompsocthanksitem P. Marsza{\l}ek is also affiliated with Doctoral School of Exact and Natural Sciences, Jagiellonian University, Kraków, Poland \protect \\
    E-mail: \{v.dubovik, p.marszalek, jacek.tabor, t.kusmierczyk\}@uj.edu.pl
    \IEEEcompsocthanksitem ${\dagger}$ Corresponding author.
}%
}


\markboth{}%
{}

\maketitle

\begin{abstract}
Low-Rank Adaptation (LoRA) enables parameter-efficient fine-tuning of large models by decomposing weight updates into low-rank matrices, significantly reducing storage and computational overhead. While effective, standard LoRA lacks mechanisms for uncertainty quantification, leading to overconfident and poorly calibrated models. Bayesian variants of LoRA address this limitation, but at the cost of a significantly increased number of trainable parameters, partially offsetting the original efficiency gains. Additionally, these models are harder to train and may suffer from unstable convergence.  
In this work, we propose a novel framework for parameter-efficient Bayesian fine-tuning, demonstrating that effective uncertainty quantification can be achieved in very low-dimensional parameter spaces. The proposed method achieves strong performance with improved calibration and generalization while maintaining computational efficiency. Our empirical findings show that, with the appropriate projection of the weight space uncertainty can be effectively modeled in a low-dimensional space, and weight covariances exhibit low ranks.
\end{abstract}


\section{Introduction}

Parameter-efficient fine-tuning methods have become a practical alternative to full fine-tuning of large pretrained models, as they substantially reduce computational and memory requirements while preserving downstream performance. Among these methods, LoRA (Low-Rank Adaptation)~\cite{hu2021lora} decomposes weight updates into low-rank matrices, enabling efficient adaptation to new tasks with only a small number of trainable parameters. 
Minimizing the number of trainable parameters reduces memory and storage requirements, making large-scale model adaptation feasible. Reducing computational overhead speeds up training time and makes adaptation possible in resource-constrained settings.  

While pretrained models are often relatively well-calibrated~\citep{openai2023gpt4}, fine-tuned large models and particularly large language models, tend to become overconfident and poorly calibrated~\citep{jiang2021can,tian2023just,xiao2022uncertainty,he2023preserving}, especially when adapted on limited data.  This hinders their usability for applications where uncertainty-aware decisions are performed. 

\begin{figure}[t]
    \centering
    \includegraphics[width=0.9\linewidth]{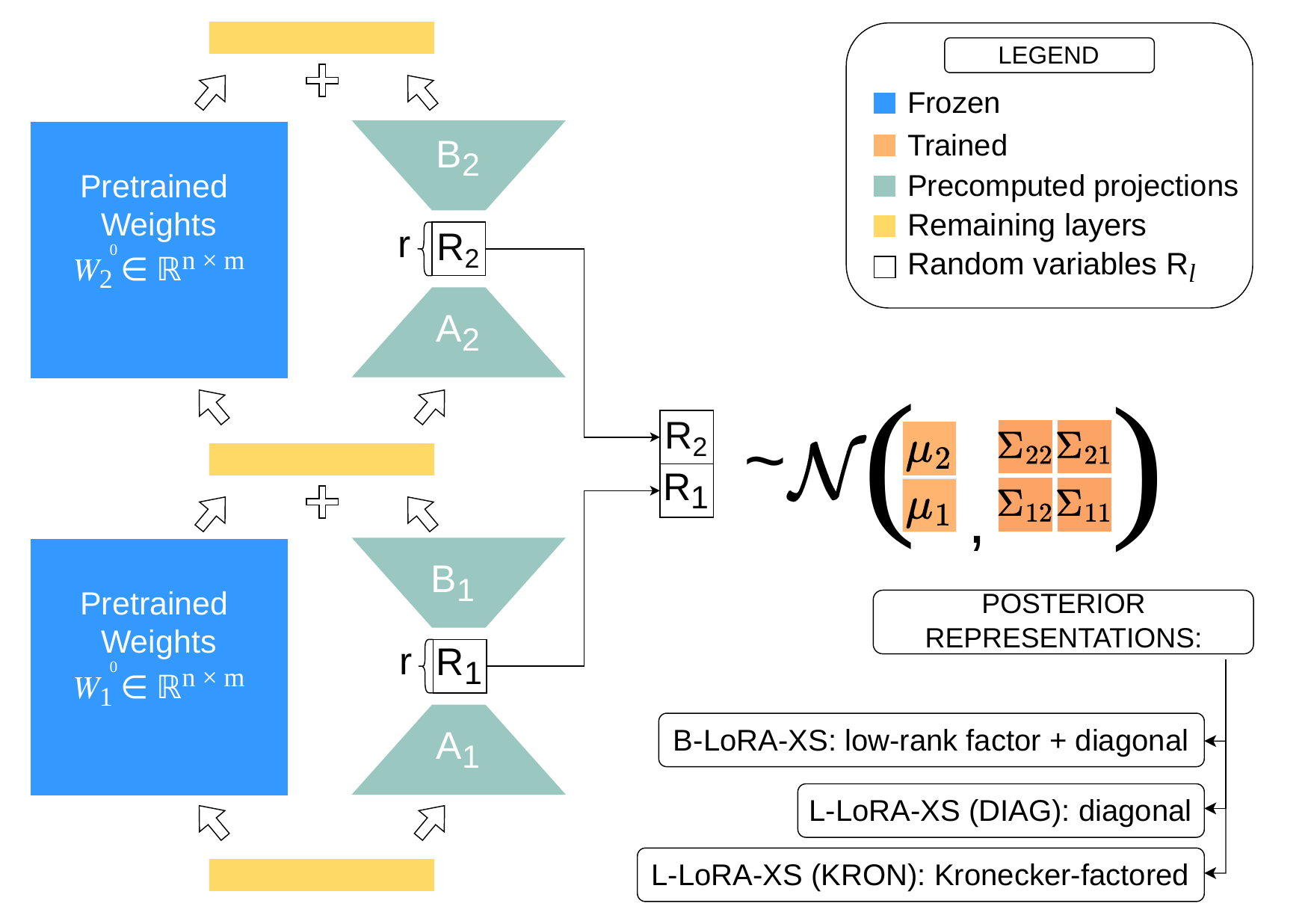}
    \caption{
    Weight adaptation using Bayesian fine-tuning in a projected subspace: original weights  (\emph{blue}) remain frozen, projections (\emph{green}) are precomputed based on the pretrained weights, and posterior parameters (\emph{orange}) are learned.
    }
    \label{fig:method}
\end{figure}

Bayesian treatment of neural networks is therefore frequently employed to mitigate overconfidence and improve uncertainty quantification~\citep{blundell2015weight,kristiadi2020being,aitchison2021deep,izmailov2021bayesian}. Consequently, recently proposed Bayesian variants of LoRA~\citep{onal2024gaussian,robeyns2024laplaceLora,doan2025bayesianlowranklearningbella} address the aforementioned challenges by introducing uncertainty estimation directly into the fine-tuning process. During training, these models jointly learn both the mean and covariance of the adapted parameters, aiming to achieve better generalization and uncertainty quantification.  

Learning the posterior covariance matrix is necessary for modeling epistemic uncertainty. However, its size grows quadratically with the number of parameters, which can easily cancel out the benefits of LoRA, in addition to making learning significantly harder. Using low-rank, Kronecker-factored, or diagonal-only covariances partially alleviates the problem, but 
this comes at the cost of results quality loss. Furthermore, even at rank $r= 2$, the number of trainable parameters is quadrupled compared to vanilla LoRA. This creates a need for an alternative approach that retains covariance modeling capacity while reducing the number of required parameters.  

We propose a framework for Bayesian fine-tuning of large pretrained models that learns Bayesian posteriors for weight updates projected onto a low-dimensional manifold, hence maintaining parameter efficiency. 
Through carefully designed projections, the framework enables learning weight updates within low-dimensional subspaces rather than in the full parameter space. 
In practice, we consider multiple instantiations of projection designs and evaluate two (currently dominant) methods for computing approximate Bayesian posteriors. 

Operating in this reduced parameter space substantially improves the feasibility of Bayesian inference. 
We show that this approach remains effective in 
capturing dependencies between parameters through structured covariances in the projected subspaces. 
In particular, we empirically demonstrate that correlations between weights can be represented efficiently in these subspaces, and unlike for the original parameterization, covariance matrices of rank as low as $k=2$ are often sufficient. 
The dimensionality reduction further stabilizes the optimization, leading to improved calibration and accuracy under limited computational budgets.

Performing Bayesian fine-tuning directly in the subspaces induced by projections from pretrained weights is both novel and technically challenging. 
Our approach enables uncertainty-aware fine-tuning in a highly parameter-efficient manner, addressing a challenge not previously explored to this extent in the literature
Although the components of the framework (projections and Bayesian methods) rely on prior work, their synergistic application to learn Bayesian posteriors within compressed subspaces represents a conceptual innovation. 

This paper is a significant extension of the work presented at EMNLP2025~\cite{marszalek-etal-2025-minimal}. While the preliminary study introduced a method for applying Bayesian inference within a subspace generated by SVD, this work presents a generalized framework for parameter-efficient Bayesian fine-tuning. A major concrete improvement is the introduction and evaluation of four distinct projection strategies, including a data-aware projection. We also employ an alternative approach for modeling Bayesian posteriors by adding the Laplace Approximation to the previously used SWAG method. Finally, the extension provides stronger theoretical grounding by linking projection-based Bayesian fine-tuning to multilinear algebra and discussing uncertainty quantification within the Bayesian framework.

Empirically, the experimental scope has been greatly expanded to evaluate the impact of the newly proposed projection and learning strategies. The evaluation has been scaled to also include a larger language model (LLaMA2-7B). 
Furthermore, the ablation studies have been extended to include a study of uncertainty distributions and decompositions, in addition to analyses of covariance matrix models and a detailed trade-off analysis between predictive accuracy and calibration metrics.

These modifications required a significant rewrite of the manuscript. In particular, we extracted the Background into a separate section (\ref{sec:background}), followed by a discussion of the general Framework in Section~\ref{sec:method}, an explanation of the projection strategies in Section~\ref{sec:projections_description}, and a discussion of approximate Bayesian learning techniques in Section~\ref{sec:bayesian_learning}. Section~\ref{sec:experiments} covers the expanded experiments, and finally, we conclude with Related Work and a discussion in Sections~\ref{sec:related_work} and \ref{sec:conclusion}.

Our implementation used to produce experimental results in this manuscript has been made publicly available~\footnote{\url{https://github.com/gmum/bayesian-finetuning}}.

\section{Background}
\label{sec:background}

Training a full deep neural network in a high-dimensional parameter space from scratch is expensive, 
as for a given weight matrix $W \in \mathbb{R}^{n \times m}$, standard training involves optimizing $n \cdot m$ entries. 
It is more effective to learn low-rank updates $\Delta W$ to a subset of frozen weights $W^0$ in a pretrained model, as the geometry of optimized loss landscapes often exhibits intrinsic dimensionality that is significantly lower than the ambient parameter space. This motivates the LoRA approach.

\textbf{LoRA} fine-tunes large pretrained models by learning low-rank weight updates $\Delta W$ instead of directly training the weights $W$. For a pretrained parameter matrix $W^0 \in \mathbb{R}^{n \times m}$, it learns a rank-$r$ update $\Delta W = AB$, where $A \in \mathbb{R}^{n \times r}$ and $B \in \mathbb{R}^{r \times m}$ have significantly fewer parameters. The effective weight is then:
$
W = W^0 + \Delta W = W^0 + AB,
$
where only $A$ and $B$ are trained. 
Then,
to achieve sufficient fidelity of fine tuning, LoRA typically is applied jointly at multiple layers, yielding a set of updates $\{\Delta W_{\ell}\}$.

\textbf{Bayesian learning} is another framework to improve the performance and calibration of machine learning models.
{Bayesian treatment} of a neural network entails computing the posterior distribution $p(\theta \mid \mathcal{D})$ given training data $\mathcal{D}$. By Bayes' theorem:
$
p(\theta \mid \mathcal{D}) = \frac{p(\mathcal{D} \mid \theta) p(\theta)}{p(\mathcal{D})},
$
where $\theta$ represents the model parameters (i.e., weights), treated as random variables, and the model likelihood is given by $p(\mathcal{D} \mid \theta) = \prod_{(x_i, y_i) \in \mathcal{D}} p(y_i \mid x_i, \theta)$.
The learned posterior enables Bayesian model averaging during inference:
\begin{align*}
p(y \mid x , \mathcal{D}) &= \int p(y \mid x , \theta) p(\theta \mid \mathcal{D}) \, d\theta \\
&\approx \frac{1}{S} \sum_{j=1}^{S} p(y \mid x , \theta_j), \quad \theta_j \sim p(\theta \mid \mathcal{D}),
\end{align*}
which improves calibration and uncertainty quantification. $S$ denotes the number of samples drawn from the posterior.

The Bayesian framework enables more extensive handling of uncertainty.
In particular,
the entropy $H[Y \mid x, \mathcal{D}]$ of the predictive distribution $p(y \mid x, \mathcal{D})$, which encodes the uncertainty of the model's predictions, by the \emph{law of total entropy}, can be decomposed \citep{NEURIPS2018_3ea2db50} for Bayesian models as:
$$
H[Y \mid x, \mathcal{D}]
  =
  \underbrace{\mathbb{E}_{\theta \sim p(\theta \mid \mathcal{D})}
                \bigl[ H[Y \mid x, \theta] \bigr]}_{\textit{aleatoric}}
  +
  \underbrace{I[Y; \theta \mid x, \mathcal{D}]}_{\textit{epistemic}},
$$
where $H[\cdot]$ denotes Shannon entropy and $I[\cdot ; \cdot]$ denotes mutual information.
The first term captures the \emph{aleatoric} (data-intrinsic) noise, 
whereas the second term measures the \emph{epistemic} (model) uncertainty, e.g., the expected information gain about $\theta$ that is obtained upon observing $y$.

Bayesian variants of LoRA (\textbf{Bayesian LoRAs}) replace point estimates of the low-rank updates with a posterior distribution over the adapted parameters 
$\theta = \bigcup_{\ell}\{A_{\ell}, B_{\ell}\}$, where $\ell$ indexes the adapted layers. 
Since the exact posterior $p(\theta \mid \mathcal{D})$ is intractable, practical approaches~\cite{robeyns2024laplaceLora,onal2024gaussian,doan2025bayesianlowranklearningbella} rely on approximations based either on particle methods or on parametric variational families. A widely adopted choice is a multivariate Gaussian approximation $p(\theta \mid \mathcal{D}) \approx \mathcal{N}(\mu, \Sigma)$, where $\mu$ represents the posterior mean and $\Sigma$ 
encodes uncertainty and statistical dependencies. 
However, the covariance matrix $\Sigma$ scales quadratically with the number of adapted parameters, making its estimation and storage computationally and memory expensive. Consequently, prior works rely on low-rank  approximations or Kronecker decompositions. However, even with these approximations, Bayesian learning for LoRA remains costly, motivating our work, which seeks to further reduce computation by compressing the learning space.

\section{Framework}
\label{sec:method}

We propose a framework for parameter-efficient Bayesian fine-tuning that combines the efficiency of LoRA with the dimensionality-reduction properties of subspace inference for Bayesian deep learning.  
Crucially, our contribution is not simply to apply Bayesian methods to LoRA, but to perform so within a compressed parameter space, thereby making Bayesian learning and inference significantly more efficient.

Let $W^0 \in \mathbb{R}^{n \times m}$ denote the pretrained weight matrix, which remains frozen throughout training. We decompose the update $\Delta W$ using two low-rank bottleneck basis matrices $A \in \mathbb{R}^{n \times r}$ and $B \in \mathbb{R}^{r \times m}$ together with a learnable core matrix $R \in \mathbb{R}^{r \times r}$. The resulting fine-tuned weight matrix is 
$$
    W = W^0 + \Delta W = W^0 + A R B .
$$
Crucially, in contrast to standard LoRA, where both factor matrices are optimized, we by default \emph{freeze} $A$ and $B$ and learn only the small core matrix $R$.

By projecting the weight updates onto the low-dimensional subspace defined by $A$ and $B$, 
we reduce the Bayesian inference problem to a tractable scale. 
The number of trainable parameters is reduced to $r^2$ per weight matrix, and since the rank $r$ is chosen such that $r \ll \min(n, m)$, it follows that $r^2 \ll (n + m) r$, resulting in a compressed parameter space.
This allows us to estimate the posterior more efficiently, avoiding the prohibitive computational costs associated with high-dimensional Bayesian learning.
Furthermore, 
we expect learning in the projected subspace to be more effective, rendering the computation of the Bayesian posterior more tractable.

The framework unifies concepts from tensor decomposition, adapter-based fine-tuning, and
subspace learning.

In multilinear algebra, our formulation is structurally identical to a \textbf{Tucker-2 decomposition} of a 2-mode tensor (matrix). In the standard Tucker model, a tensor $W$ is decomposed into a core tensor multiplied by a factor matrix along each mode.
However, while standard Tucker decomposition algorithms (like Higher-Order SVD~\cite{doi:10.1137/S0895479896305696}) seek to update $A$, $B$, and $R$ simultaneously to approximate $W$, our framework treats $A$ and $B$ as static inductive biases, 
isolating the learning dynamics within the core $R$.

In the context of \textbf{Parameter-Efficient Fine-Tuning}, our method is closely related to LoRA, but differs in both structure and trainability. First, we adopt a "sandwich" parameterization $A R B$ rather than a simple low-rank product $A\cdot B$. Second, whereas LoRA learns the basis matrices $A$ and $B$, we instead fix these bases and learn only the interaction matrix $R$. This is a \textit{spectral adapter}, where optimization is performed in the spectral domain defined by $A$ and $B$.

Finally, our method is a 
 specific instance of \textbf{Subspace Inference} framework (SI)~\citep{izmailov2020subspace}, which was proposed as a remedy for the intractability of full-dimensional posteriors in modern networks. {Low-dimensional affine manifold}, carefully centred on a well-trained solution, already contains a rich family of high-performing weight vectors. Performing Bayesian integration restricted to that manifold restores calibrated uncertainty without revisiting the entire parameter space.
 
In partiuclar,
given a well-trained reference point $\bar w\in\mathbb{R}^{d}$ and $k\ll d$ orthonormal basis vectors (e.g., the leading PCA directions of an SGD trajectory) the \emph{learning subspace} is
$$
\mathcal{S}:=\bigl\{ w=\bar w+Pz  \bigm|  z\in\mathbb{R}^{k}\bigr\},\qquad P\in\mathbb{R}^{d\times k}.
$$
where Bayesian parameters are the low-dimensional representations $\theta = \{z\}$.

Whereas SI \emph{learns} $P$ from the training dynamics, we \emph{project} onto directions already favored by the pretrained backbone (see Section~\ref{sec:projections}). 
We define a \emph{layer-local} manifold that leaves the pretrained backbone untouched. 
In particular, for each frozen weight matrix $W^{0}_{\ell}$, 
fixed projectors $A_{\ell}$ and $B_{\ell}$ are  computed once.
Then, 
all task-specific variability is captured by these square adapters $R_{\ell}\in\mathbb{R}^{r\times r}$.

Vectorising and stacking these projections defines
$$
w = \bar{w}+P_{\tB} z_{\tB}, \qquad
P_{\tB}=\mathrm{blockdiag}\bigl(B_{\ell}^{\top}\otimes A_{\ell}\bigr),
$$
where $w = [ \mathrm{vec}(W_1)^T,  \ldots,  \mathrm{vec}(W_\ell)^T,  \ldots]^T$, $\bar{w} = [ \mathrm{vec}(W_1^0)^T,  \ldots,  \mathrm{vec}(W_\ell^0)^T,  \ldots]^T$, and $P_{\tB}$ is a block-diagonal matrix with blocks $B_{\ell}^{\top} \otimes A_{\ell}$, one for each layer~$\ell$.

Thus our method explores an affine subspace 
$$
\mathcal{S}_{\tB}:=\{w = w^{0}+P_{\tB}z_{\tB}| z_{\tB}\in\mathbb{R}^{\sum_{\ell} r^{2}}\},
$$
 whose dimension scales with $r^{2}$ per adapted layer.

\section{Projections}
\label{sec:projections_description}

The representational power of the proposed approximation $\Delta W \approx A R B$ is fundamentally constrained by the subspaces spanned by $A$ and $B$. As these bases remain frozen during the learning phase, their initialization imposes a strong inductive bias, effectively defining the 'vocabulary' available to the core matrix $R$. The optimization process is thus reduced to finding the optimal mixing coefficients within this predefined geometric structure.
Consequently, the selection of $A$ and $B$ is a design choice that determines the trade-off between compression, computational efficiency, and adaptation capability.

We consider four distinct approaches to computing the projection matrices $A$ and $B$: Truncated SVD for task-agnostic weight reconstruction, the data-aware Truncated Whitened SVD minimizing activation errors, a 2D Discrete Cosine Transform for frequency-domain modeling via universal bases, and structure-agnostic Random Projections using Gaussian orthonormal vectors.

For brevity, whenever we discuss projecting a single weights matrix $W_\ell$, we omit the layer index $\ell$, and simply write $W$. However, in practice to increase modeling flexibility the projections are applied to multiple layers $\{\ell\}$ at the same time.

\subsection{Truncated SVD}
Our default approach builds on the assumption that the target fine-tuning task is similar to the original task on which the model was pretrained. Consequently, the adaptation projections $A$ and $B$ should be constructed to align with the principal directions spanned by the original weight matrix $W^0$.
Formally, the objective is specified as minimization of the Frobenius norm of the reconstruction error $\mathcal{L}=\|W^0 - \hat{W}\|_F$ subject to the low-rank constraint $\hat{W} = A R B$ (with $R=I$).
The optimal solution is obtained via the truncated SVD of $W^0$, by setting $A = U_r \Sigma_r$ and $B = V_r^T$, following the design of LoRA-XS~\cite{balazy2024lora}.
This decomposition of $W^0$ ensures that the frozen bases capture the most informative singular components of the original weights.

The default SVD-based approach serves as the theoretical baseline for matrix compression, focusing solely on minimizing the reconstruction error of the weights.
By minimizing the Frobenius norm, it identifies the principal components of the weight matrix, i.e., directions corresponding to the largest singular values, and discards the remainder.
However,
while mathematically optimal, 
this method is agnostic to the neural network's actual function.
In particular, small weights that are crucial due to their interaction with highly active inputs may be discarded, potentially degrading model performance.

\subsection{Truncated Whitened SVD}
An alternative approach, based on {whitened SVD}~\cite{yuan2023asvd,wang2024svd}, addresses the limitations of the default projection by accounting for the statistical distribution of the input data 
(in the context of fine-tuning, this refers to the training dataset used for a task), e.g., covariance $\Sigma_{xx}$. 
Instead of preserving the weights directly, it aims to preserve the layer's output, minimizing activation errors rather than weight errors. By whitening the inputs, this method effectively scales the importance of weights based on the variance of the features they multiply.
This potentially yields superior accuracy, especially for layers with outlier features, but comes at the cost 
of estimating the covariance matrix $\Sigma_{xx}$.

Formally, we seek an approximation $\hat{W} = A R B$ of rank $r$ that minimizes the expected squared error:
$
\mathcal{L} = \mathbb{E}_x \left[ \| xW^0 - x\hat{W} \|^2 \right].
$
This expands to a weighted Frobenius norm involving the input covariance matrix $\Sigma_{xx}$ as 
$
\mathcal{L} = \| (W^0 - \hat{W})^T \Sigma_{xx}^{1/2} \|_F^2.
$

The optimal solution is obtained via the SVD of the {whitened weight matrix}
$
\tilde{W} = (W^0)^T \Sigma_{xx}^{1/2} \approx \tilde{U}_r \tilde{S}_r \tilde{V}_r^T.
$
This effectively projects the weights onto the principal components of the input data.
The {forward whitener} transforms raw inputs to a decorrelated (white) space as $P = \Sigma_{xx}^{1/2} = Q \Lambda^{1/2} Q^T$, where the eigendecomposition of the covariance is given by $\Sigma_{xx} = Q \Lambda Q^T$. The {inverse whitener} $P^{-1} = \Sigma_{xx}^{-1/2} = Q \Lambda^{-1/2} Q^T$ transforms the vectors from the white space back to the raw input space.

The output basis $B$ corresponds directly to the left singular vectors $\tilde{U}_r$. These vectors map to the output dimension $m$, yielding:
$
B = \tilde{U}_r^T.
$
Constructing the input basis $A$ is more challenging. The right singular vectors $\tilde{V}_r$ exist in the whitened space. To apply them to raw inputs, we must apply the inverse whitener $P^{-1}$ to compute the raw projection vectors $V_{raw} = P^{-1} \tilde{V}_r = \Sigma_{xx}^{-1/2} \tilde{V}_r$.
While the analytical solution for the core matrix is simply the singular values $\tilde{S}_r$, we incorporate these into $A$ (as for $R=I$) leading to the final form of 
$
A = \Sigma_{xx}^{-1/2} \tilde{V}_r \tilde{S}_r.
$

Finally, let's note that we can calculate the covariance matrices $\Sigma_{xx}$ very efficiently using Welford's Algorithm~\cite{welford1962note,chan1983algorithms}. 
We use the incremental online algorithm on batched data to maintain a constant (per layer) memory footprint regardless of dataset size.

\subsection{2D Discrete Cosine Transform}
DCT-II~\cite{ahmed1974discrete}
differs fundamentally from the above SVD-based approaches
by using a fixed, universal basis of cosine waves. 
It approximates the weight matrix $W^0$ by transforming it to the frequency domain, truncating low-energy coefficients, and projecting using the selected basis vectors.

Formally, the orthogonal transformation matrix $D \in \mathbb{R}^{d \times d}$ is defined by row indices $k$ (frequency) and column indices $i$ (spatial) as
$ D_{k,i} = \alpha_k \cos\left(\frac{\pi k (2i + 1)}{2d}\right), $
where the normalization factor $ \alpha_k = \begin{cases} \sqrt{\frac{1}{d}} & \text{if } k=0 \\ \sqrt{\frac{2}{d}} & \text{if } k > 0 \end{cases}$ ensures orthogonality (i.e., $D^T = D^{-1}$).
The pretrained weights $W^0$ are transformed into the frequency coefficient matrix $C$ using the row-transform matrix $D_n \in \mathbb{R}^{n \times n}$ and column-transform matrix $D_m \in \mathbb{R}^{m \times m}$ as
$ C = D_n W^0 D_m^T  \in \mathbb{R}^{n \times m} $. 

To maximize reconstruction quality under a fixed rank constraint, we select the frequency components maximizing the signal energy. By Parseval's theorem~\cite{oppenheim1999discrete}, the total energy in the spatial domain is preserved in the frequency domain, i.e., $\|W^0\|_F^2 = \|C\|_F^2$. Consequently, minimizing the Frobenius norm of the reconstruction error is equivalent to retaining the coefficients with the largest squared magnitudes.
We compute the marginal energy for each row frequency $i$ and column frequency $j$ as:
$   \mathcal{E}_i^{\text{row}} = \sum_{j=1}^{m} C_{i,j}^2$ {and} $ \mathcal{E}_j^{\text{col}} = \sum_{i=1}^{n} C_{i,j}^2$.
The index sets $\mathcal{I}$ and $\mathcal{J}$ are then constructed by selecting the top-$r$ indices corresponding to the largest values in respectively rows and columns. This ensures that the truncation retains the dominant spectral bands of the weight matrix.
The \emph{core matrix} 
$ C_{\mathcal{I}, \mathcal{J}} \in \mathbb{R}^{r \times r} $
is the submatrix of coefficients at these intersections.

{The left projection basis $A$ is formed by taking the columns of the inverse row transform ($D_n^T$) corresponding to the selected row frequencies $\mathcal{I}$. Then, we integrate in it also the core matrix to eventually get 
${A} = (D_n^T)_{:, \mathcal{I}} \cdot C_{\mathcal{I}, \mathcal{J}}$ (for $R=I$). 

The right projection basis $B$ is formed by taking the rows of the column transform ($D_m$) corresponding to the selected column frequencies $\mathcal{J}$ leading to
$ {B} = (D_m)_{\mathcal{J}, :} $.

Standard DCT relies on the input signal being locally smooth to concentrate energy in low-frequency coefficients~\cite{rao2014discrete}. However, neural network weight matrices are typically unstructured and stochastic with high-frequency variations.
To mitigate this, we introduce a deterministic \textit{global permutation} step. We define permutation matrices $P_r \in \{0,1\}^{n \times n}$ and $P_c \in \{0,1\}^{m \times m}$ that reorder the rows and columns of $W^0$ monotonically based on their $L_1$ norms. The compression is then performed on the sorted matrix $\tilde{W} = P_r W^0 P_c^T$ instead.
This reordering clusters weights of similar magnitude and transforms the unstructured weight landscape into a quasi-monotonic surface. This artificially induces spatial correlation, shifting signal energy into fewer low-frequency coefficients and significantly improving the approximation capability of the DCT.
Crucially, these permutations do not introduce inference-time overhead. The inverse permutations are absorbed directly into the projection factors $\tilde{A}$ and $\tilde{B}$ derived from $\tilde{W}$:
$
    W^0 \approx P_r^T (\tilde{A} \tilde{B}) P_c = (P_r^T \tilde{A}) (\tilde{B} P_c) = A B.
$
Thus, the final projectors $A$ and $B$ restore the original connectivity without requiring explicit index mapping during the forward pass.

\subsection{Random Projections}

DCT assumes the weight matrix has  smooth structure, effectively acting as a low-frequency filter. As an alternative we consider random projections~\cite{doi:10.1137/090771806}, which make no structural assumptions 
regarding weight matrices.

Formally, to generate the left projector, we sample a Gaussian matrix $G_L \in \mathbb{R}^{n \times r}$ with i.i.d. entries $G_{ij} \sim \mathcal{N}(0,1)$ and compute its reduced QR decomposition $G_L = Q_L R_L$. Haar uniformity on the Stiefel manifold~\cite{doi:10.1137/090771806,doi:10.1137/0717034,james1964distributions} is enforced by fixing the signs of the diagonal of $R_L$, yielding
$
L = Q_L \cdot \operatorname{diag}(\operatorname{sgn}(\operatorname{diag}(R_L))),
$
so that $L^\top L = I_r$ and $L$ is Haar-distributed on $V_r(\mathbb{R}^n)$.
The use of Haar measure ensures that the projectors are uniformly distributed among all orthogonal $n \times r$ and $r \times m$ matrices, respectively.
 
Analogously, the right projector is obtained by sampling $G_R \in \mathbb{R}^{m \times r}$ with i.i.d. Gaussian entries, computing $G_R = Q_R R_R$, and correcting signs to obtain
$R_{\text{temp}} = Q_R \cdot \operatorname{diag}(\operatorname{sgn}(\operatorname{diag}(R_R))),
$
which satisfies $R_{\text{temp}}^\top R_{\text{temp}} = I_r$ and is Haar-distributed on $V_r(\mathbb{R}^m)$. The right factor is then set as $B = R_{\text{temp}}^\top$.

The core interaction matrix is obtained by projecting $W^0$ onto these random subspaces,
$C = L^\top W^0 R_{\text{temp}} \in \mathbb{R}^{r \times r},
$
and the final left factor is $A = L C$.

This procedure exploits concentration-of-measure phenomena akin to the Johnson-Lindenstrauss lemma~\cite{ledoux2001concentration}: with high probability, random orthogonal projections preserve the essential spectral structure of $W^0$ in the reduced space while avoiding the cost of data-dependent eigen-decompositions. 

\section{Bayesian Fine-Tuning in Projected Subspaces}
\label{sec:bayesian_learning}

Bayesian fine-tuning seeks a posterior distribution 
$p(\{\Delta W_{\ell}\}\mid\mathcal{D})$ over the weight updates.
However,
thanks to 
the subspace projections  
$\Delta W_{\ell}=A_{\ell}R_{\ell}B_{\ell}$ this posterior can be expressed via the posterior over the compressed matrices $\{R_{\ell}\}$ instead. 
In particular,
we infer a joint posterior over the inner parameters  
$
p(\theta=\bigcup_{\ell}R_{\ell}\mid\mathcal{D})\approx\mathcal{N}(\mu,\Sigma),
$
where $\Sigma$ may capture both intra-layer and cross-layer dependencies.

Although we never form the full covariance of $\Delta W_{\ell}$ explicitly, it is implicitly defined via the following Kronecker structure and could in principle be computed as
\begin{equation}
\Sigma_{\Delta W_{\ell}}
=(B_{\ell}^{T}\!\otimes A_{\ell})\,\Sigma_{R_{\ell}}\,(B_{\ell}^{T}\!\otimes A_{\ell})^{T},
\end{equation}
where $\Sigma_{R_{\ell}}$ denotes the layer-specific covariance over $R_{\ell}$ and $\otimes$ is the Kronecker product.  

During inference, we sample $R_{\ell}$ from the learned posterior and reconstruct the corresponding updates yielding $\Delta W_{\ell}$ from $A_{\ell}$,$R_{\ell}$,$B_{\ell}$, without ever materializing full weight matrices or their covariances.

The parameters $\mu$ and $\Sigma$ can be learned using various Bayesian approaches. We focus on two efficient and effective methods: SWAG~\cite{NEURIPS2019_118921ef} and the Laplace Approximation~\cite{ritter2018scalable}. 

\subsection{SWAG}

SWAG~\cite{NEURIPS2019_118921ef} constructs a Gaussian approximation to the posterior by exploiting the trajectory of stochastic gradient descent. 
First, it converges to a region surrounding a local minimum. Then, in the second stage, 
it aggregates statistics of iterates after convergence, interpreting the variability of SGD solutions as samples from a local posterior basin. 

The procedure yields a low-rank plus diagonal covariance that captures both local curvature and non-isotropic uncertainty without explicit second-order computation.
Specifically,
after a burn-in phase (a fixed 10 or 25 epochs) of the gradient-based optimization, the algorithm maintains  a running average of $\theta$ named $\hat \mu$, along with $k$ vectors of differences 
$$\hat D_{{last}} = \theta_{{last}} - \hat \mu$$ 
for the last $k$ values of $\theta$, and a running average of $\theta^2$. Based on these averages, we estimate the variances $\hat \sigma^2$ for individual parameters and approximate the covariance as
$$
 \Sigma \approx \frac{1}{2}(\hat D \cdot \hat D^T + {diag}(\hat \sigma^2)),
$$
which constitutes a rank-$k$ approximation to the covariance matrix.
Overall, the method utilizes a total of $|\theta|\cdot (k+2)$ parameters (per module per layer), where $|\theta| = \sum_{\ell} r^2$.

\subsection{Laplace Approximation}

Laplace inference~\cite{ritter2018scalable,daxberger2021laplace} is a post-hoc Bayesian procedure that approximates the posterior around a maximum-a-posteriori (MAP) solution. In our implementation, the MAP phase follows standard LoRA fine-tuning, e.g., we optimize the updates  
$
\Delta W_\ell = A_\ell R_\ell B_\ell .
$
Unlike for SWAG,
two training regimes are possible in the MAP phase:  
(i) only the core matrices $\{R_\ell\}$ are optimized while $\{A_\ell,B_\ell\}$ are fixed; or alternatively
(ii) all matrices $\{A_\ell,B_\ell,R_\ell\}$ are optimized.

Crucially, regardless of how MAP is performed, the Bayesian step treats only the core matrices
$
\theta = \bigcup_\ell R_\ell
$
as random variables. The projection matrices $\{A_\ell,B_\ell\}$   only appear as deterministic linear maps when relating uncertainty in $R_\ell$ to uncertainty in $\Delta W_\ell$. They are frozen and do not enter 
posterior estimation.

Given $\theta_{\text{MAP}}=\bigcup_\ell R_\ell^{\text{MAP}}$, we approximate the posterior by a second-order expansion of the log-joint~\cite{antoran2022adapting} as
\[
p(\theta\mid\mathcal{D}) \approx \mathcal{N}(\theta_{\text{MAP}},\Sigma), \quad 
\Sigma = -\big(\nabla^2_{\theta}\log p(\mathcal{D}\mid \theta)+\lambda I\big)^{-1}.
\]
The primary challenge in finding the Laplace approximation of the posterior is estimating the Hessian matrix. In particular, this raises problems regarding memory storage and computational cost, as well as numerical stability issues. In practice, the Hessian is replaced by a positive-definite surrogate, such as the Fisher information matrix or the GGN.

\subsubsection{Covariance structure}

Because $\Sigma$ is too large to store explicitly even over $\{R_\ell\}$, we consider two approximations:

\paragraph{Diagonal Laplace (DIAG)}  
We keep only the marginal variances of each entry in $R_\ell$, i.e.,
$
\Sigma_{R_\ell} \approx \mathrm{diag}(\sigma^2_{R_\ell}).
$
This requires storing $r^2$ variances per layer and assumes independence both \emph{within} a layer and \emph{across} layers.  
Because $\Sigma_{R_\ell}$ is diagonal, this approximation ignores all correlations between elements of $R_\ell$ and across layers. It is memory-optimal but significantly less expressive.

\paragraph{Kronecker-factored Laplace (KFAC/KRON)}  
For each layer we approximate the Fisher of $R_\ell$ with a Kronecker block,
\[
F_\ell \approx \mathbb{E}\!\left[(a_{\ell-1}a_{\ell-1}^T)\otimes(g_\ell g_\ell^T)\right],
\]
where $a_{\ell-1}$ are layer inputs and  
$g_\ell=\nabla_{b_\ell}\log p(y\mid x,\theta)$ are output gradients.  
We store this block in a low-rank form of rank $k_{\text{kfac}}$, accumulated incrementally during training.

\subsubsection{Inter-layer covariances}  
Within a layer, correlations are modeled through the Kronecker factors.  
Across layers, we do \emph{not} build a dense full-block covariance. Instead, we share a common low-rank basis across layers by stacking the per-layer Kronecker factors and fitting a joint low-rank representation. This yields a tractable approximation that captures cross-layer dependencies without materializing a full $L r^2 \times L r^2$ matrix.

\subsubsection{Model complexity}

Let $r$ denote the LoRA rank and $L$ the number of adapted layers.  
Then, in the MAP phase, the number of optimized parameters depends on which matrices are trained. If all LoRA components $\{A_\ell,B_\ell,R_\ell\}$ are optimized, the total number of MAP parameters equals the sum over layers of the parameters in the projection matrices plus the core matrix, i.e., $\sum_\ell (r\,n_\ell + r\,m_\ell + r^2)$. If only the core matrices $\{R_\ell\}$ are trained, then the MAP phase involves exactly $\sum_\ell r^2$ ($=L \times r$) parameters.  

Regardless of the MAP strategy, the Bayesian posterior is defined exclusively over the core matrices $\{R_\ell\}$, so the dimensionality of the posterior space is always $\sum_\ell r^2$.  

The representation of the covariance further determines the storage cost. With a diagonal Laplace approximation, only the marginal variances of each entry in $R_\ell$ are kept, requiring storage of only $\sum_\ell r^2$ numbers. 
In contrast, a Kronecker-factored Laplace approximation scales as $O(L\, r^4)$ if full factors are computed (our case), or alternatively,  as $O(L\, r^2\, k_{\text{kfac}})$ parameters with $k_{\text{kfac}} < r^2$ whenever low-rank factors are used.
This retains rich structure while remaining tractable.

Overall, MAP training may involve $\{A_\ell,B_\ell,R_\ell\}$, but Bayesian computation scales only with the core $r^2$ parameters per layer. Diagonal Laplace is cheapest but ignores correlations while Kronecker Laplace is slightly heavier but captures both intra-layer and controlled cross-layer dependencies while remaining tractable.

Using Laplace Approximation for deep neural networks requires linearization of model around $\theta_{\text{MAP}}$~\cite{daxberger2021laplace,antoran2022adapting} as
$
f_\theta(x^*) \approx f_{\theta_{\text{MAP}}}(x^*) 
+ J(x^*)^\top(\theta-\theta_{\text{MAP}}),
$
which induces a Gaussian over logits with covariance  
$\Lambda = J\,\Sigma\,J^\top$.  
Samples of $\theta$ (or equivalently logits) are drawn using the low-rank representation but full weight matrices or full covariances are never materialized.

\section{Experiments}
\label{sec:experiments}

Experiments presented in this section compare the effectiveness of different projections (Section~\ref{sec:projections}), evaluate the utility and trade-offs of our approach compared to the baselines for RoBERTa (Section~\ref{sec:performance_roberta}) and LLaMA (Section~\ref{sec:performance_llama}), explore uncertainty and OOD detection (Section~\ref{sec:experiments_ood}), and illustrate ablation of 
covariance matrix ranks for B-LoRA-XS model (Section~\ref{sec:cov_matrix_rank_analysis}).
The ablation of  robustness against data size reduction can be found in the Supplement (Section~S2).

\subsection{Setup} 

In the experiments, we evaluate three variants of our proposed approach: 
(1) {B-LoRA-XS (B-XS)}, trained using SWAG~\cite{NEURIPS2019_118921ef}; 
(2) {L-LoRA-XS (L-XS)}, trained using the Laplace approximation~\citep{ritter2018scalable,daxberger2021laplace} with either a Kronecker-factored (KRON/K) or diagonal (DIAG/D) covariance matrix; and 
(3) {L-LoRA-S (L-S)}, also trained using the Laplace approximation, but where the projections $A$ and $B$ additionally were fine-tuned during the MAP phase.

We evaluated our models on four GLUE tasks~\cite{wang2019} (RTE, MRPC, CoLA, and SST-2) for RoBERTa-large~\cite{liu2019} employed as the base architecture. To assess scalability and verify whether our findings generalize to larger models, we also report experiments on LLaMA2-7B~\cite{touvron2023llama} in Section~\ref{sec:performance_llama}. 

We compare our approach against standard LoRA and its parameter-efficient variant, namely LoRA-XS. For the Bayesian baseline, we employ LoRA-SWAG~\cite{onal2024gaussian}. For the LLaMA~2-7B experiments (Section~\ref{sec:performance_llama}), we also included results copied from the Laplace-LoRA (LA) paper~\cite{robeyns2024laplaceLora}.

For the more compact models (i.e., LoRA-XS and our variants), we studied ranks $r \in [2, 25]$ for RoBERTa and up to $100$ for LLaMA. For LoRA and LoRA-SWAG, we tested $r \in \{2, 8\}$ due to computational constraints. The total number of parameters  (which we use as a proxy for storage and computational cost) as a function of rank $r$ (and rank $k$ for SWAG or KRON/DIAG for Laplace) is summarized in Fig.~\ref{fig:lorar}~(RoBERTa) and Table~\ref{tab:LLaMA_params}~(LLaMA).

We report accuracy (higher is better), ECE, and NLL (lower is better).
Accuracy reflects the quality of fit, while ECE and NLL serve as proxies for calibration and uncertainty. 

Further details of the setup are provided in the Supplement

\subsection{Subspace Projection Impact}
\label{sec:projections}

\begin{figure}
    \centering
    \includegraphics[width=1.0\linewidth]{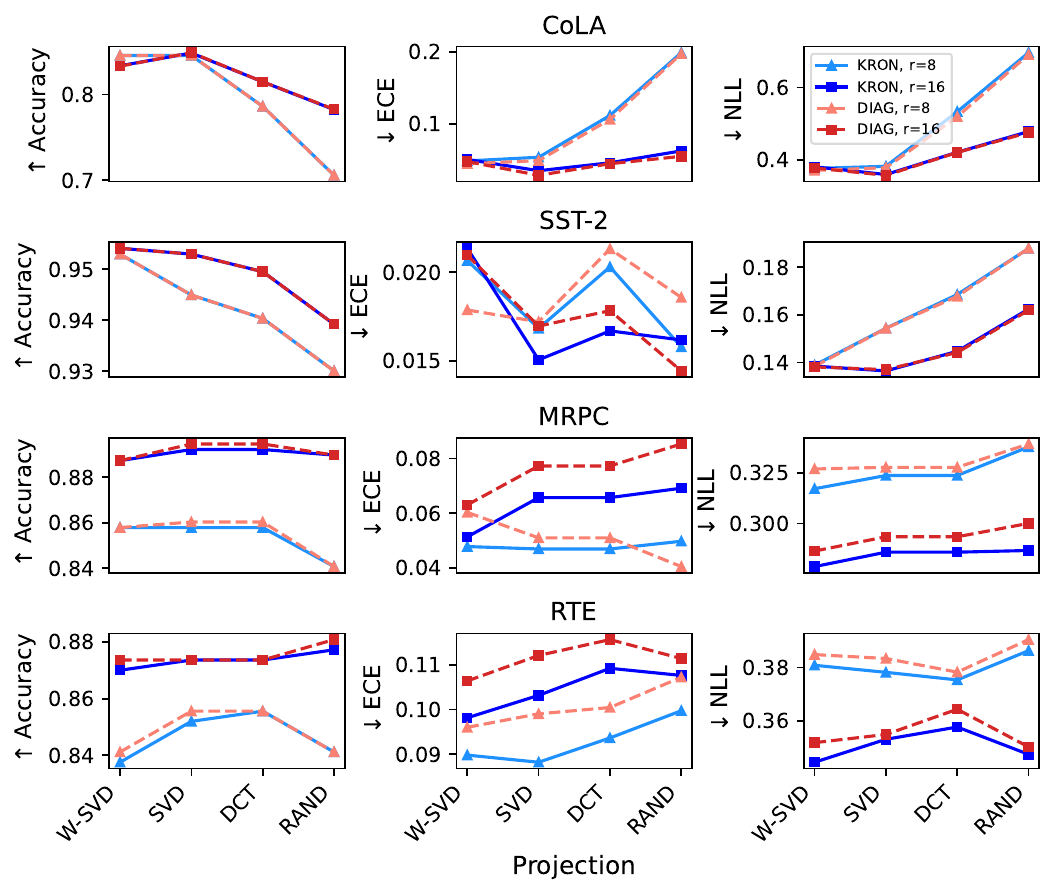}
    \caption{Impact of projection choice on Accuracy, ECE, and NLL for Laplace-based fine-tuning of RoBERTa-Large. We compare four projections (W-SVD, SVD, DCT, RAND) for two ranks ($r=8,16$) and two covariance models (DIAG=red lines and KRON=blue lines). 
    The exact numerical values we report in Table~S1 
    in the Supplement.
    }
    \label{fig:results_roberta_projections}
\end{figure}

\begin{figure}
    \centering
    \includegraphics[width=1.0\linewidth]{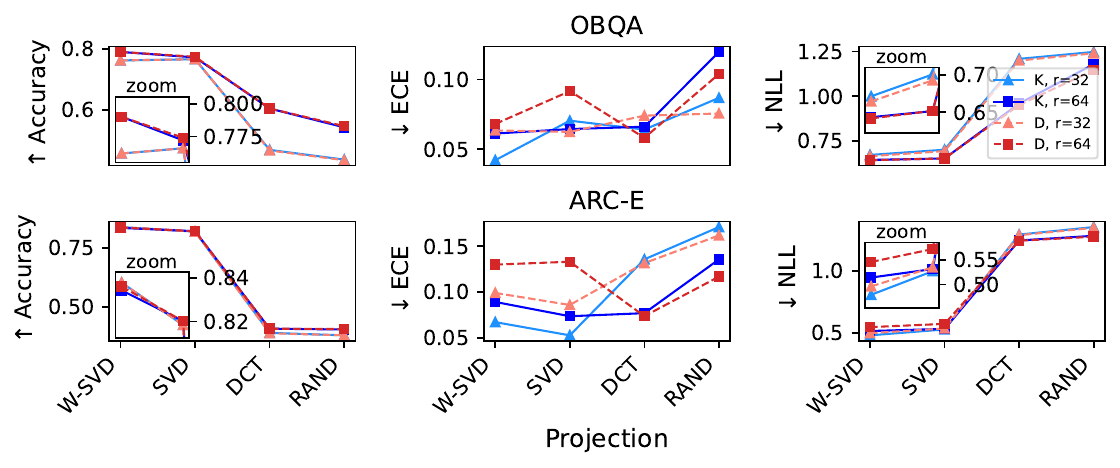}
    \caption{Impact of projection choice on Accuracy, ECE, and NLL for Laplace-based fine-tuning of LLaMA2-7B. We compare four projections (W-SVD, SVD, DCT, RAND) for two ranks ($r=32,64$) and two covariance models (DIAG=red lines and KRON=blue lines). 
    The exact numerical values we report in Table~S2 
    in the Supplement.
    }
    \label{fig:results_llama_projections}
\end{figure}

Fig.~\ref{fig:results_roberta_projections} presents the effect of varying projections by comparing four alternatives (Section~\ref{sec:projections_description}):  W-SVD, standard SVD, DCT, and random (RAND). We tested RoBERTa-Large with both diagonal (DIAG) and Kronecker (KRON) Laplace covariances at ranks $r \in \{8, 16\}$, employed across four tasks. Table~S1
in the Supplement reports additional results for hybrid projections combining two bases (WSVD+SVD, DCT+SVD, and RAND+SVD), where half of the subspace is constructed using one approach and the other half using SVD.

The study disentangles three design aspects: (i) the projection, (ii) the rank $r$, and (iii) the covariance structure (DIAG vs.\ KRON). The clearest and most consistent patterns emerge on the two largest datasets, namely CoLA and SST-2 (top two). For most cases, W-SVD yields the highest Accuracy and lowest NLL, indicating that whitening produces a subspace that best aligns with functionally relevant directions. However, the standard SVD is almost on par with W-SVD and in some cases may overtake it, which we attribute to noisy behavior due to random initialization.
CoLA ECE follows the same clear ordering as Accuracy and NLL.
On the other hand,
on SST-2, ECE values are very small for all methods, which makes ECE differences  noisy and non-informative. In this regime (=when ECE is near-zero) NLL provides a more reliable comparison. Regardless of ECE, we conclude that both \textbf{SVD variants are still clearly superior to the remaining projections}. Overall, DCT performs moderately, while RAND suffers a substantial degradation, particularly at $r=8$, where accuracy and NLL deteriorate dramatically.

On the smaller datasets (MRPC and RTE), trends are qualitatively similar but less pronounced and more noisy. In particular, for RAND, depending on the initialization, we were able to obtain surprisingly good results. Overall, Accuracy varies little across projections, yet SVD and especially W-SVD still tend to dominate in NLL. For RTE, ECE is sufficiently large to reveal the pattern of the dominance of the SVD-based methods, while for MRPC, ECE differences are more modest and noisy, for example for DIAG with rank $r=8$ we observe a behavior opposite to the the general trend.

Comparing covariance models, KRON generally yields slightly better NLL and ECE than DIAG, especially for weaker projections (DCT and RAND), but it does not alter the ranking of the methods (i.e., projections). 
This suggests that while \textbf{structured covariance helps, it cannot compensate for a poor subspace}. On the other hand, \textbf{varying the rank $r$ from 8 to 16 noticeably improves results for all projections}. Gains from increasing $r$ are significant, yet for CoLA and SST-2, they are insufficient to impact the relative ordering of the projections, for example, Accuracy for DCT with $r=16$ is still worse than for SVD with $r=8$.

\begin{figure*}[ht!]
\centering
\begin{minipage}{0.7\textwidth}
    \includegraphics[width=1.0\linewidth]{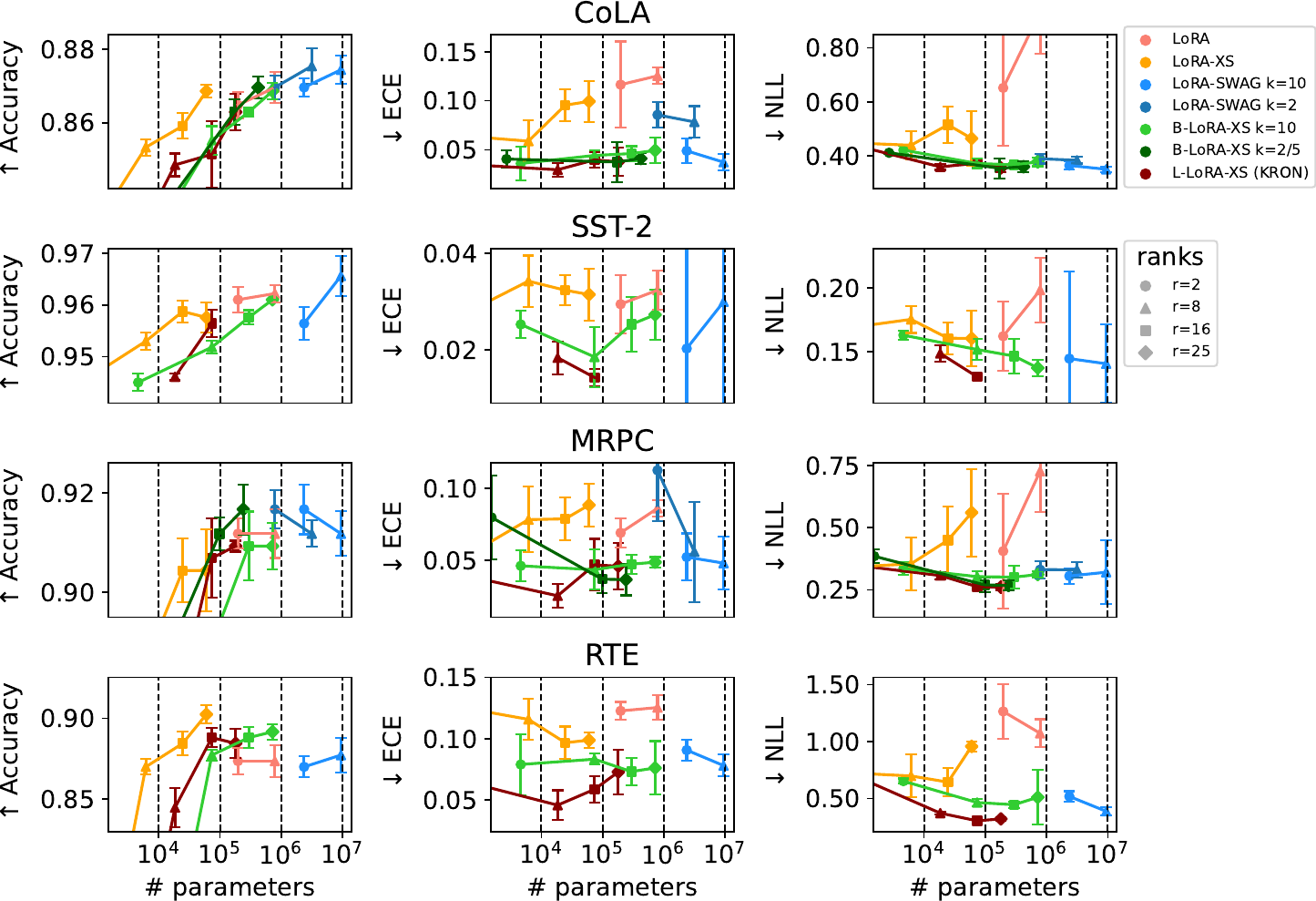}
\end{minipage}    
\hfill
\begin{minipage}{0.28\textwidth}
    \centering
    \resizebox{1.0\linewidth}{!}{
        \begin{tabular}{llrrr}
            \toprule
             & \textbf{Method} & \textbf{$r$} & $k$ & \textbf{\# Params} \\
            \cmidrule{1-5}
            \multirow{4}{*}{\rotatebox{90}{MAP}} & LoRA & 2 & - & 0.2M \\
            & LoRA & 8 & - & 0.8M \\
            \cmidrule{2-5}
            & LoRA-XS & 8 & - & 6k \\
            & LoRA-XS & 25 & - & 60k \\
            \midrule
            \multirow{10}{*}{\rotatebox{90}{Bayesian}} & LoRA-SWAG & 2 & 10 & 2.4M \\
            & LoRA-SWAG & 8 & 10 & 9.4M \\
            & LoRA-SWAG & 8 & 5 & 5.5M \\
            \cmidrule{2-5}
            & {\our} & 8 & 10 & 74k \\
            & {\our} & 25 & 10 & 0.7M \\
            & {\our} & 25 & 5 & 0.4M \\
            \cmidrule{2-5}
            & L-LoRA-XS & 8 & KRON & 18k \\
            & L-LoRA-XS & 16 & KRON & 73k \\
            & L-LoRA-XS & 25 & KRON & 0.2M \\
            \bottomrule
        \end{tabular}
    }
\end{minipage}%
\caption{        
    Median$\pm$std. accuracy (left), ECE (middle), and NLL (right) on 4 GLUE tasks (rows) vs. total parameter count$^{\star}$ for several methods and varying ranks~$r$. \our{} and L-LoRA-XS (ours) achieve the accuracy and the calibration of LoRA-SWAG (a standard Bayesian approach) while using significantly fewer parameters than LoRA (the default deterministic variant).
    \textcolor{black}{
    The exact numerical values underlying the plots we report in Tables~S3-S6 
    in the Supplement.
    }
    {${}^{\star}$ For Bayesian approaches, we report the number of parameters used for computing the Gaussian posterior.}
}
\label{fig:lorar}
\end{figure*}

The hybrid approaches (W-SVD+SVD, DCT+SVD, and RAND+SVD) reported in Table~S1 
in the Supplement provide further insight into the role of projection basis diversity. Mixing a weaker, data-agnostic basis with SVD (DCT+SVD and RAND+SVD) partially recovers performance (especially in NLL) relative to using pure DCT or RAND, but remains consistently inferior to full SVD and W-SVD across all datasets. In contrast, W-SVD+SVD performs significantly worse than either basis alone on CoLA and offers no clear benefit elsewhere. This happens because W-SVD and SVD span highly correlated directions, and combining them yields highly correlated basis vectors. This effectively reduces the usable dimensionality of the projection (e.g., $r=16$ has capacity close to $r=8$). We conclude that our \textbf{hybrids never outperform the separate projections}, indicating that simply mixing bases is insufficient unless the added directions are both complementary and data-aligned.

Fig.~\ref{fig:results_llama_projections} repeats the above analysis for the LLaMA2-7B model. The patterns for the OBQA and ARC-E datasets mirror those observed previously for CoLA and SST-2. Regarding Accuracy and NLL, 
a distinct gap remains between SVD-based approaches
and the remaining two projections. \textbf{W-SVD marginally improves over the vanilla SVD projeciton, at the cost of precomputing $\Sigma_{xx}$}. In contrast, ECE results lack clear patterns, with behavior varying across different ranks and covariance types.

Together, these results support the central observation that effective \textbf{Bayesian fine-tuning depends critically on learning in a well-chosen low-dimensional subspace}. In particular, {W-SVD} and SVD define manifolds that preserve task-relevant structure while enabling tractable posterior inference. In contrast, more generic projections and naive hybrids substantially constrain what the Bayesian posterior can represent, regardless of the covariance model.

\subsection{Trade-offs Under Parameter Budgets} 
\label{sec:performance_roberta}

Fig.~\ref{fig:lorar} reports accuracy, ECE, and NLL as a function of the total number of parameters for the RoBERTa-Large architecture. We compare several classes of methods: deterministic low-rank adaptations (LoRA and LoRA-XS) and a Bayesian extension (LoRA-SWAG) against our subspace Bayesian approaches, which include B-LoRA-XS (the SWAG variant) and for completeness, also L-LoRA-XS (the Laplace-based variant).
We omit the Laplace-LoRA baseline~\cite{robeyns2024laplaceLora} due to differences in preprocessing (following~\cite{balazy2024lora}), which make the results not directly comparable.
%
Finally, throughout this section, we focus on the SVD-based projection as a representative and computationally efficient choice.

The experiment was designed to study the trade-off between predictive performance, calibration quality, and model complexity. We report the total number of parameters on the horizontal axis as a unified measure of model cost.
Reporting in this way enables a consistent comparison across methods and 
isolates the measurements from implementation-specific effects. 
Note that for the Bayesian variants this denotes the number of parameters used in the second stage of training for fitting the posterior. 

Our main observation is that the proposed \textbf{subspace Bayesian approaches improve overall model performance, with a particular emphasis on calibration}. Fig.~\ref{fig:lorar} (middle and right) shows that both B-LoRA-XS and L-LoRA-XS consistently achieve lower ECE and NLL than standard LoRA across all parameter scales. While the two variants follow similar trends, L-LoRA-XS often matches the calibration performance of B-LoRA-XS with fewer posterior parameters, reflecting the efficiency of the Laplace approximation. 

Regarding accuracy (Fig.~\ref{fig:lorar}, left), standard LoRA attains marginally better performance for a few configurations at intermediate parameter counts. However, \textbf{in the majority of settings our approaches match or exceed the accuracy of standard LoRA}. Importantly, we do not observe any setting in which standard LoRA outperforms the Bayesian variants in terms of calibration, which is central for this work.

More broadly, Bayesian methods-including B-LoRA-XS, L-LoRA-XS, and LoRA-SWAG consistently outperform their non-Bayesian counterparts in ECE and NLL. Compared to LoRA-SWAG, \textbf{the subspace Bayesian approaches achieve similar or better calibration quality while using substantially fewer parameters}, typically by a factor of 5-15. Between the two our variants, B-LoRA-XS tends to provide slightly stronger calibration at higher parameter budgets, but L-LoRA-XS demonstrates comparable performance for most regimes. 

Moreover, while LoRA-SWAG can perform well in some configurations, its results vary noticeably across parameter scales and covariance ranks. In contrast, both B-LoRA-XS and L-LoRA-XS exhibit stable and consistent behavior. This is particularly evident on MRPC and CoLA, where performance remains robust across different values of $k$, whereas LoRA-SWAG's ECE degrades at lower ranks (e.g., $k=2$).

\subsection{Uncertainty and Out-of-Distribution Detection}
\label{sec:experiments_ood}

\begin{table}
\caption{
OOD predictive entropies for models fine-tuned on CoLA.
}
    \centering
    {
\begin{tabular}{l|rr}
\toprule
  & \multicolumn{2}{c}{$H  \bigl[Y\mid x,\mathcal D\bigr]$} \\
 method & MRPC & SST-2 \\
\midrule
  B-LoRA-XS & $0.35_{0.15}$ & $0.42_{0.11}$ \\
  LoRA-SWAG & $0.36_{0.06}$ & $0.43_{0.04}$ \\
  LoRA-XS & $0.12_{0.15}$ & $0.23_{0.17}$ \\
 LoRA & $0.02_{0.02}$ & $0.05_{0.03}$ \\
\bottomrule
\end{tabular}
}
\label{tab:entropy}
\end{table}

\begin{figure}
    \centering
    \begin{minipage}{0.65\linewidth}
    \includegraphics[width=0.49\linewidth]{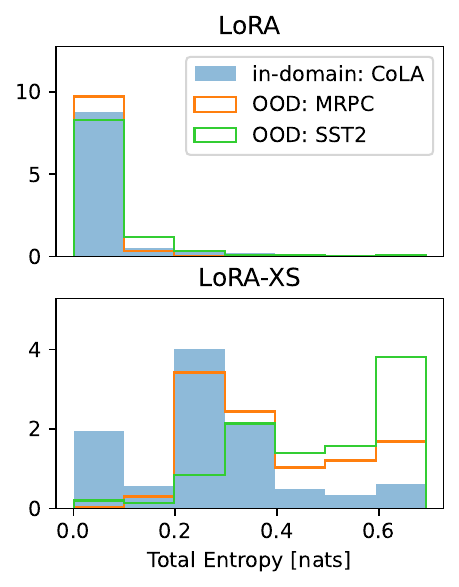}
    \includegraphics[width=0.475\linewidth]{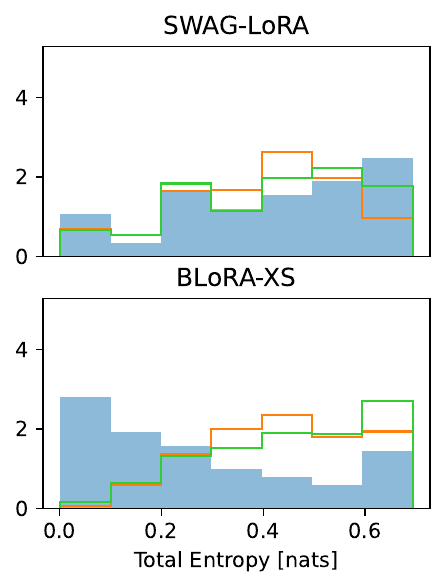}        
    \end{minipage}
  \caption{
Predictive uncertainty distributions for in-domain and OOD:
histograms of predictive entropy for models trained on CoLA and evaluated on in-domain CoLA test data (blue) and two OOD datasets, MRPC (orange) and SST-2 (green). The left column shows results for deterministic baselines, whereas the right column shows values for their Bayesian variants. 
}
    \label{fig:uncertainty_scores}
\end{figure}

\begin{table}
\caption{
OOD detection performance using uncertainty-based scores.
}
    \centering
    {
\begin{tabular}{l|lrrrr}
\toprule
  & \multicolumn{2}{c}{MRPC} & \multicolumn{2}{c}{SST-2}  \\
method  & $W_1$ & AUROC & $W_1$ & AUROC \\
\midrule
B-LoRA-XS (total) & 0.16 & 0.73 
 & 0.18 & 0.74 \\
LoRA-SWAG (total) & 0.05 & 0.44 
 & 0.02 & 0.48 \\
LoRA-XS (total) & 0.13 & 0.70 
 & 0.23 & 0.84 \\
LoRA (total) & 0.04 & 0.67 
 & 0.03 & 0.75 \\
 \midrule
 B-LoRA-XS (epistemic) & 0.04 & 0.58
 & 0.05 & 0.64 \\
LoRA-SWAG (epistemic) & 0.04 & 0.43
 & 0.03 & 0.45 \\
\bottomrule
\end{tabular}
}
\label{tab:entropy_auroc}
\end{table}

\begin{figure}
    \centering
    \includegraphics[width=0.65\linewidth]{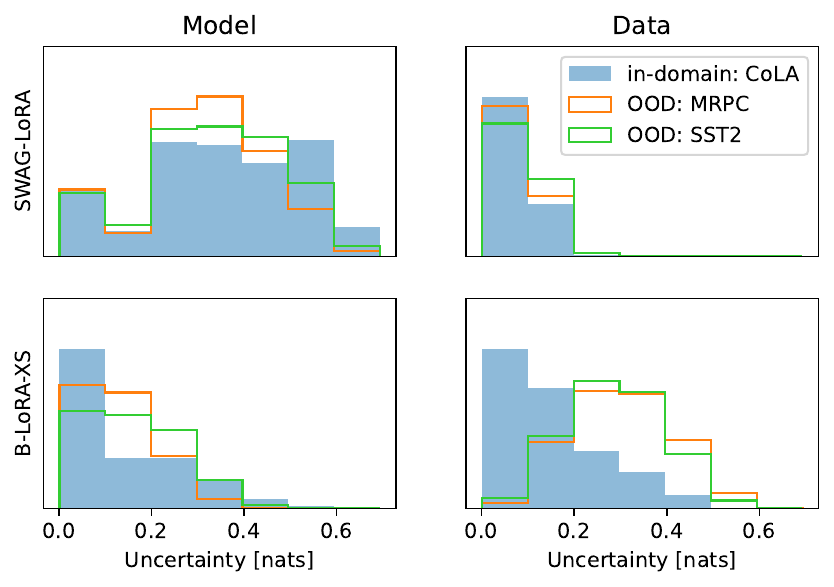}
  \caption{
Decomposition of predictive uncertainty into epistemic (model) and aleatoric (data) uncertainty for B-LoRA-XS and LoRA-SWAG, evaluated on in-domain (CoLA) and OOD datasets (MRPC and SST-2). 
}
    \label{fig:uncertainty_scores_decomposition}
\end{figure}

To evaluate the capability of the trained models to distinguish between in-domain (ID) and out-of-distribution (OOD) data, we analyzed their predictive entropy. The results are summarized in Table~\ref{tab:entropy}.
The table presents average predictive entropy values and standard deviations computed across multiple random seeds for models trained on CoLA and tested on two OOD datasets: MRPC and SST-2.
\textbf{Bayesian variants, namely B-LoRA-XS and LoRA-SWAG, yield significantly higher entropy scores compared to their deterministic counterparts}, LoRA and LoRA-XS, whose predictive entropies are notably lower and, in the case of standard LoRA, often close to zero, pointing toward general overconfidence.

Fig.~\ref{fig:uncertainty_scores} illustrates these results through histograms of predictive entropy. We note significant differences between models. LoRA scores concentrate in regions of high confidence regardless if the data is in-domain or OOD, indicating overconfident predictions. LoRA-XS and LoRA-SWAG exhibit more variability in predictive entropy. However, B-LoRA-XS, which combines Bayesian learning with projection-based adaptation, exhibits the most favorable behavior. \textbf{For B-LoRA-XS, predictive entropy scores for in-domain data concentrate on the left-hand side of the plot, whereas scores for OOD data are shifted toward higher values on the right-hand side}.

In Table~\ref{tab:entropy_auroc}, we measure how these properties translate to the practical ability to detect OOD data using total predictive- and also epistemic entropy. The results show that total predictive entropy provides a stronger separation between in-domain and OOD samples. In particular, \textbf{the results show the effectiveness of B-LoRA-XS and also LoRA-XS in distinguishing ID from OOD data}, as evidenced by higher Wasserstein-1 distances~\citep{villani2008optimal} ($W_1$ between $0.13$ and $0.23$) and AUROC values~\citep{FAWCETT2006861} ($0.70$ to $0.84$). Conversely, standard LoRA and LoRA-SWAG exhibit weaker discrimination capabilities, with relatively lower Wasserstein-1 distances and AUROC scores, indicating their limited reliability for OOD detection.

Our results suggest that \textbf{OOD detection capabilities come from two sources: Bayesian learning and learning in projected subspaces}, which improves the models' detection ability through the aleatoric component of predictive uncertainty.
%
In particular,
Fig.~\ref{fig:uncertainty_scores_decomposition} exhibits discrepancies between uncertainty distributions for in-domain and OOD data. While epistemic (left-hand side) uncertainty already shows significant differences between ID and OOD data, the largest separation for B-LoRA-XS is observed in the aleatoric (right hand-side) component. 
Moreover, for this model the uncertainty scores follow a clear ordering, with OOD samples generally assigned higher uncertainty than in-domain samples.

\subsection{Covariance Matrix Rank Analysis} 
\label{sec:cov_matrix_rank_analysis}

Fig.~\ref{fig:results_roberta_projections}~and~\ref{fig:results_llama_projections} illustrate the difference between using KRON vs. DIAG covariances for L-LoRA-XS. 
An analogous study for B-LoRA-XS is presented in Fig.~\ref{fig:covrank}.
In particular,
the figure compares the sensitivity of the Bayesian LoRA variants to changes in covariance matrix rank $k$.  
Markers indicate model sizes (e.g., LoRA-SWAG $\gg$ \our).

\begin{figure}[t!]
    \begin{minipage}{1.0\linewidth}
    \centering
    {\scriptsize 
    \hspace{0.8cm} CoLA\\
    \vspace{0.025cm}
    \includegraphics[width=1.0\linewidth]{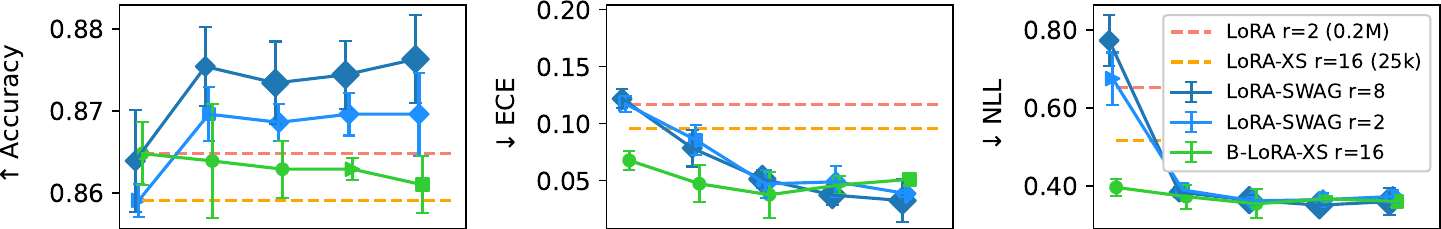} \\
    \vspace{0.01cm}
    \hspace{0.8cm} MRPC\\
    \vspace{0.025cm}
    \includegraphics[width=1.0\linewidth]{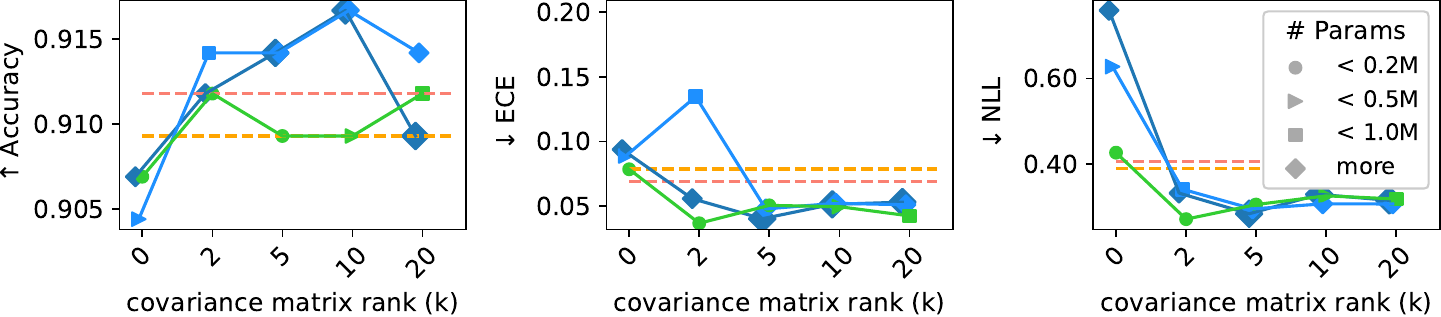}\\
    }    
    \end{minipage}
    \caption{
    Impact of the posterior covariance matrix rank ($k=0$ indicates the case with no off-diagonal elements) for CoLA (top) and MRPC (bottom). For brevity, confidence bars ($\pm$ standard deviation) are omitted for MRPC. The colored lines represent non-Bayesian baselines (e.g., standard LoRA or LoRA-XS at a given rank $r$).   
    \textcolor{black}{    The exact numerical values underlying the plots are reported in Tables~S7~and~S8 
    in the Supplement.}    
}
    \label{fig:covrank}
\end{figure}

As expected, LoRA-SWAG deteriorates proportionally as rank decreases. On the other hand, \textbf{\our{} maintains its performance across a wide range of $k$}. Significant degradation occurs only when off-diagonal covariance values are entirely ignored (i.e., at $k=0$). Notably, \our{} achieves the best calibration at low ranks, particularly at $k=2$ or $k=5$. This demonstrates that the SVD-based projection effectively disentangles parameters, enabling low-dimensional uncertainty modeling.

\subsection{Performance Trade-Offs on a Larger Model}
\label{sec:performance_llama}

To evaluate our framework's scalability and efficacy on larger architectures, we fine-tuned LLaMA2-7B on three commonsense reasoning datasets. 
In particular, we compare our approach, in both Laplace- and SWAG-based (XS) variants using SVD-based projections, against the standard \underline{Laplace (LA)} approximation baseline. LA represents the state-of-the-art for Bayesian LoRA and has been shown to outperform SWAG on this architecture~\cite{robeyns2024laplaceLora}.
%
%
Additionally, we consider a modified variant of our method (L-S) using Laplace-based training, but with the projection matrices $A$ and $B$ also fine-tuned during the MAP fitting phase. Specifically, in the first phase we perform LoRA-like fine-tuning with a slightly increased budget (by $L \times r^2$ additional parameters to fine-tune also the $R$ matrices), whereas in the 2nd phase we run the standard posterior fitting. This strategy trades computational efficiency for increased modeling capacity of the projections. 

Table~\ref{tab:LLaMA_params} summarizes the number of parameters used by each of the methods at each training stage.  
Given the parameter efficiency of our approaches, we could increase for them the number of modules to which fine-tuning was applied while still being more efficient than- or on par with the baselines.  
The details 
are discussed in the Supplement.

Table~\ref{table:LLaMA_results} demonstrates that \textbf{transitioning from the MAP estimate to the Bayesian posterior consistently yields significant improvements in calibration and uncertainty} (ECE and NLL) across all methods, with minimal impact on Accuracy. The critical difference between the methods is the parameter efficiency with which these calibration gains are achieved.
In particular,
while the Laplace baseline (LA) achieves strong calibration, it does so by relying on an expensive Bayesian posterior. At the same time, our methods achieve comparable, and occasionally superior, calibration with drastically fewer Bayesian parameters. 

L-LoRA-S (L-S; 10M variant), which uses the number of parameters comparable to the LA baseline, delivers a clear accuracy improvement over LA across all three tasks (e.g., $80.6\%$ vs.\ $78.9\%$ on OBQA) while matching or surpassing its calibration (e.g., reaching an ECE of $4.6$ vs.\ $5.4$ on ARC-E and an NLL of $0.6$ vs.\ $0.65$ on OBQA).
This confirms that the \textbf{computational budget can be spent more effectively than it is for the standard Bayesian LoRA baselines}.
On the other hand, \textbf{at a fraction of the parameter cost, the compressed variants of our approach (i.e., XS variants) also deliver significant calibration benefits}. For instance, L-LoRA-XS ($r=64$) utilizes less than a third of the budget of the Laplace baseline while sharply reducing ECE (for example, ECE=$7.9$ on ARC-C). Although the absolute accuracy of these heavily compressed models is marginally lower than that achieved by the top approaches, it remains highly competitive on ARC-E and OBQA, offering an unparalleled trade-off for memory-constrained regimes.
We furthermore observe that \textbf{increasing the rank (from $32$ to $100$) steadily improves accuracy, whereas the calibration benefits saturate relatively early}.

Table~S9 in the Supplement shows that the \textbf{results from Table~\ref{table:LLaMA_results} can be further improved through more careful hyperparameter selection}. In particular, selecting an appropriate checkpoint from the MAP training trajectory of the model (instead of the fixed checkpoint used here) for Bayesian posterior estimation yields further improvements across all metrics. For example, the ECE for OBQA at rank $r=32$ decreases from $6.0$ to $4.3$, while for ARC-E from $6.6$ down to $5.5$.

The calibration and accuracy gains observed for models trained with SWAG are less pronounced than for those using the Laplace approximation, despite both methods receiving nearly identical treatment during the MAP phase (Laplace is trained until convergence to a local minimum, whereas SWAG only until convergence to a \emph{region} around the minimum). This performance gap is particularly evident regarding ECE and NLL for
the OBQA and ARC-C datasets. Despite both strategies have similar expressiveness, the former achieves significantly better fits. This highlights the \textbf{importance of selecting robust Bayesian learning techniques for large-scale and complex models}.

Overall, our results confirm that \textbf{uncertainty can be effectively captured in low-dimensional spaces}, provided the weight space is appropriately projected and Bayesian posteriors learned in a robust way.

\begin{table}[t]
\centering
\caption{Number of parameters 
used at each training stage.}
\resizebox{1.0\linewidth}{!}
{%
\begin{tabular}{l|rrrr}
\toprule
{Method} & {1st stage (MAP)} & {2nd stage (Bayesian)} & {Total} \\
\midrule
\underline{Laplace (LA)}  & 4.5M & 5.6M & 11.1M \\
L-S  (4M, r=8)  & 4.2M & 8.1k & 4.2M \\
L-S  (10M, r=8)  & 10.1M & 16k & 10.1M \\
L-XS (r=100)  & 2.2M & 4.5M & 6.7M \\
L-XS (r=64)  & 0.9M & 1.8M & 2.7M \\
L-XS (r=32)  & 0.2M & 0.4M & 0.6M \\
B-XS (r=64, k=5)  & 0.9M & 6.4M & 7.3M \\
B-XS (r=32, k=5)  & 0.2M & 1.6M & 1.8M \\
\bottomrule
\end{tabular}
\label{tab:LLaMA_params}
}
\end{table}

\begin{table}[t]
\small
\centering
\caption{
Accuracy and calibration of 
MAP vs. Bayesian fine-tuning of LLaMA2-7B across three commonsense reasoning tasks.
}
\label{table:LLaMA_results}
\resizebox{1.0\linewidth}{!}
{
\begin{tabular}{l|l|cc|cc|cc}
    \toprule
    \multirow{2}{*}{} & \multirow{2}{*}{Method} & \multicolumn{2}{c|}{OBQA} & \multicolumn{2}{c|}{ARC-E} & \multicolumn{2}{c}{ARC-C} \\
    &     & MAP    & Bayesian    & MAP    & Bayesian    & MAP    & Bayesian \\
    \midrule
    \multirow{7}{*}{\rotatebox{90}{Accuracy $\uparrow$}} \rule{0pt}{2.25ex}
    & \underline{Laplace (LA)}    & $78.7_{0.4}$    & $78.9_{0.2}$    & $84.7_{1.5}$    & $85.1_{1.5}$    & $66.3_{0.6}$    & $65.3_{0.2}$ \\
    \rule{0pt}{2.25ex}
    & L-S  (10M)    & $80.8_{0.9}$    & $80.6_{1.1}$    & $86.4_{0.6}$    & $85.9_{0.3}$    & $69.8_{2.8}$    & $68.8_{2.3}$ \\
    & L-S  (4M)    & $78.8_{0.1}$    & $78.8_{0.2}$    & $84.3_{0.2}$    & $84.5_{0.3}$    & $64.4_{1.9}$    & $65.4_{1.3}$ \\
    \rule{0pt}{2.25ex}
    & L-XS r=100    & $80.8_{1.2}$    & $80.2_{1.3}$    & $84.0_{0.8}$    & $84.0_{1.2}$    & $63.1_{0.9}$    & $63.1_{1.2}$ \\

    & L-XS r=64    & $76.8_{0.3}$    & $78.0_{0.3}$    & $83.2_{0.4}$    & $82.7_{0.7}$    & $58.3_{1.5}$    & $61.0_{1.4}$ \\
    & L-XS r=32    & $76.0_{1.6}$    & $75.4_{1.3}$    & $82.9_{1.2}$    & $82.7_{0.8}$    & $52.2_{3.6}$    & $53.6_{3.3}$ \\
    
    & B-XS r=64    & $78.2_{0.6}$    & $78.6_{0.9}$    & $82.9_{0.3}$    & $83.4_{0.4}$    & $61.8_{3.2}$    & $62.3_{2.2}$ \\
    & B-XS r=32    & $76.8_{0.7}$    & $76.4_{0.6}$    & $82.2_{0.2}$    & $82.4_{0.3}$    & $52.5_{4.7}$    & $54.2_{5.1}$ \\
     
    \midrule
    \multirow{7}{*}{\rotatebox{90}{ECE $\downarrow$}} \rule{0pt}{2.25ex}
    & \underline{Laplace (LA)}    & $16.1_{0.6}$    & ${6.4_{0.8}}$    & $13.4_{1.3}$    & ${5.4_{0.2}}$    & $31.0_{0.5}$    & ${7.4_{0.7}}$ \\
    \rule{0pt}{2.25ex}
    & L-S  (10M)    & $16.4_{1.4}$    & $6.5_{1.7}$    & $11.8_{0.5}$    & $4.6_{1.0}$    & $27.1_{2.1}$    & $6.4_{2.6}$ \\
    & L-S  (4M)    & $16.5_{0.5}$    & $10.1_{0.2}$    & $12.4_{0.8}$    & $4.6_{1.0}$    & $21.9_{1.7}$    & $10.0_{0.5}$ \\
    \rule{0pt}{2.25ex}
    & L-XS r=100    & $15.6_{1.5}$    & $10.7_{0.9}$    & $12.3_{1.1}$    & $7.9_{2.4}$    & $23.9_{0.7}$    & $9.6_{0.6}$ \\

    & L-XS r=64    & $18.0_{0.4}$    & $9.2_{1.3}$    & $10.6_{1.1}$    & $8.5_{0.4}$    & $23.3_{4.8}$    & $7.9_{0.3}$ \\
    & L-XS r=32    & $18.5_{1.8}$    & $6.0_{0.8}$    & $9.2_{1.4}$    & $6.6_{0.5}$    & $19.4_{2.9}$    & $6.0_{0.7}$ \\
    
    & B-XS r=64    & $14.4_{0.7}$    & $10.7_{0.7}$    & $14.1_{0.5}$    & $5.1_{0.6}$    & $29.9_{3.4}$    & $20.5_{1.2}$ \\
    & B-XS r=32    & $15.7_{1.0}$    & $10.2_{1.1}$    & $12.8_{0.6}$    & $6.3_{0.6}$    & $29.3_{1.6}$    & $24.6_{0.5}$ \\

    \midrule
    \multirow{7}{*}{\rotatebox{90}{NLL $\downarrow$}} \rule{0pt}{2.25ex}
    & \underline{Laplace (LA)}  & $0.99_{0.05}$    & ${0.65_{0.01}}$    & $1.26_{0.13}$    & ${0.49_{0.06}}$    & $3.28_{0.29}$    & ${0.88_{0.03}}$ \\
    \rule{0pt}{2.25ex}
    & L-S  (10M)    & $1.12_{0.20}$    & $0.60_{0.02}$    & $0.76_{0.06}$    & $0.46_{0.01}$    & $1.87_{0.14}$    & $0.88_{0.02}$ \\
    & L-S  (4M)    & $1.09_{0.15}$    & $0.77_{0.05}$    & $0.76_{0.07}$    & $0.49_{0.01}$    & $1.33_{0.17}$    & $0.99_{0.06}$ \\
    \rule{0pt}{2.25ex}
    & L-XS r=100    & $1.12_{0.06}$    & $0.63_{0.01}$    & $0.79_{0.11}$    & $0.51_{0.01}$    & $1.45_{0.05}$    & $0.98_{0.02}$ \\

    & L-XS r=64    & $1.16_{0.11}$    & $0.66_{0.01}$    & $0.67_{0.07}$    & $0.53_{0.00}$    & $1.30_{0.21}$    & $0.97_{0.03}$ \\
    & L-XS r=32    & $1.05_{0.04}$    & $0.67_{0.01}$    & $0.60_{0.05}$    & $0.52_{0.01}$    & $1.26_{0.05}$    & $1.09_{0.01}$ \\
    
    & B-XS r=64    & $0.91_{0.05}$    & $0.74_{0.02}$    & $0.97_{0.03}$    & $0.60_{0.02}$    & $1.97_{0.37}$    & $1.46_{0.17}$ \\
    & B-XS r=32    & $0.96_{0.02}$    & $0.71_{0.02}$    & $0.89_{0.01}$    & $0.61_{0.02}$    & $1.72_{0.35}$    & $1.53_{0.06}$ \\
    \bottomrule
    \end{tabular}
}
\end{table}

\section{Related Work}
\label{sec:related_work}
We build on several closely related lines of prior work: parameter-efficient adaptation of pretrained models via low-rank LoRA adapters, Bayesian treatment of adapter parameters for uncertainty quantification, and restriction of Bayesian inference to a low-dimensional subspace following subspace inference. 
We discuss each of these in turn below.

\paragraph{PEFT} As large language models continue to grow, parameter-efficient fine-tuning (PEFT) has become a popular approach to reducing computational and storage costs. Among various methods~\cite{houlsby2019parameter,guo-etal-2021-parameter,li2021prefix,lester2021power}, LoRA~\cite{hu2021lora} has emerged as one of the most widely used. 
Building on its success, several approaches have been proposed to enhance different aspects of PEFT~\cite{kopiczko2023vera,adalora,dettmers2024qlora}. One such method, LoRA-XS~\cite{balazy2024lora}, further improves parameter efficiency by enabling flexible control over the number of trainable parameters per adaptation module.
We reuse the idea of SVD-based projections to reduce the dimensionality. 
However, we extend it to alternative subspaces and place it within a Bayesian framework.

\paragraph{Bayesian LoRAs} Standard LoRA~\cite{hu2021lora} does not account for uncertainty, making fine-tuned models susceptible to miscalibration. Then, Bayesian LoRA approaches integrate Bayesian inference techniques into LoRA to improve uncertainty estimation and generalization.

Several Bayesian LoRA methods have been proposed, each employing different Bayesian techniques to address these challenges.  LoRA-SWAG \cite{onal2024gaussian} combines Stochastic Weight Averaging-Gaussian (SWAG) with LoRA to enable approximate Bayesian inference, significantly improving model calibration and reducing overconfidence. Laplace-LoRA~\cite{robeyns2024laplaceLora} applies a Laplace approximation to the posterior over LoRA parameters. Bella~\cite{doan2025bayesianlowranklearningbella} introduces an approach that reduces the cost of Bayesian deep ensembles by applying multiple low-rank perturbations to a pretrained model.
BLoB (Bayesian Low-Rank Adaptation by Backpropagation) \cite{wang2024blob} jointly learns both the mean and covariance of model parameters throughout the fine-tuning process using Variational Inference. B-LoRA \cite{meo2024bayesianlora} introduces a Bayesian perspective to both quantization and rank selection by using a prior distribution over these hyperparameters, optimizing model efficiency and reducing bit operations.
The major issue with these Bayesian methods is the higher storage and memory requirements.
Although we utilize posterior learning approaches previously used by SWAG- and Laplace-LoRA, we apply them in projected subspaces, constituting respectively B-LoRA-XS and L-LoRA-XS.

\paragraph{Subspace Inference} Bayesian inference in deep networks can be made tractable by performing it in a low-dimensional subspace. This is thanks to the observation that much of the posterior mass relevant for predictions lies on a low-dimensional manifold~\cite{doi:10.1073/pnas.2310002121,izmailov2021bayesian}. 
Hence,  subspace inference methods learn posteriors in projected subspaces, enabling scalable uncertainty estimation without modeling the full parameter space \citep{izmailov2020subspace,izmailov2021bayesian}.
Some authors explore structured subspaces for Bayesian inference~\citep{daxberger2021bayesian}.
Other approaches such as SWAG~\citep{NEURIPS2019_118921ef} fit a Gaussian posterior in a low-rank (+diagonal) subspace  defined by stochastic gradient trajectories.
We follow the general principle of learning posteriors in subspaces, but differ significantly from prior work in how the subspaces are defined. In particular, performing fine-tuning instead of full training provides access to the pretrained weight matrix $W^0$, which enables subspace identification without the costly exploration of optimization trajectories.

\section{Conclusion}
\label{sec:conclusion}

This work addresses the lack of methods for principled and  parameter-efficient uncertainty quantification in LoRA. We tackle this challenge by proposing weight projections to restrict model updates to lower-dimensional subspaces while employing Bayesian framework for handling uncertainty. Under such a transformation, combining low-rank parameter updates with suitable low-rank or factorized covariance approximations achieves predictive calibration that matches other Bayesian methods, 
while utilizing a fraction of their parameters.

The primary strength of our method lies in its calibration capabilities: it consistently achieves lower expected calibration error and negative log-likelihood compared to standard LoRA and its compressed variant, LoRA-XS, across a wide range of parameter scales. Although standard LoRA may exhibit marginally better accuracy for a few specific configurations, our approaches match or exceed its accuracy in most settings and, critically, always provide improved calibration. Furthermore, compared to the Bayesian LoRA baselines, our method generally matches or surpasses their accuracy and calibration performance while using significantly fewer parameters, exhibiting greater training stability, and relying on simpler, lower-rank covariance representations.

We conclude that an appropriate weight-space transformation that combines low-dimensional parameter updates with suitable Bayesian posterior covariance approximations can improve both predictive performance and calibration or at least offer a synergistic trade-off.

\section*{Acknowledgments}

This research is part of the project No. \textbf{2022/45/P/ST6/02969}
co-funded by the National Science Centre and the European
Union Framework Programme for Research and Innovation
Horizon 2020 under the Marie Skłodowska-Curie grant
agreement No. 945339. For the purpose of Open Access, the
authors have applied a CC-BY public copyright licence to any
Author Accepted Manuscript (AAM) version arising from
this submission. 
\includegraphics[width=0.75cm]{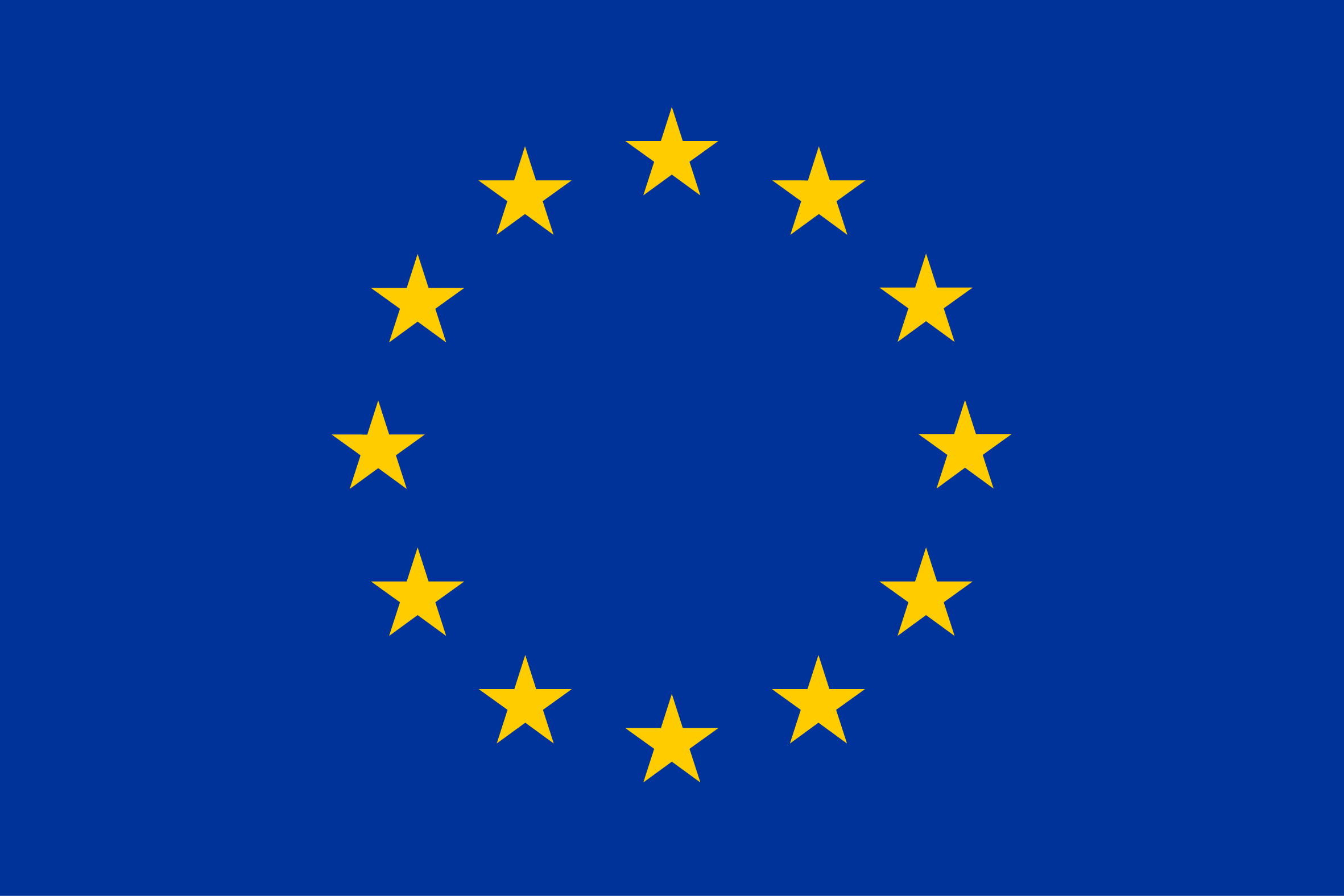} \includegraphics[width=1.475cm]{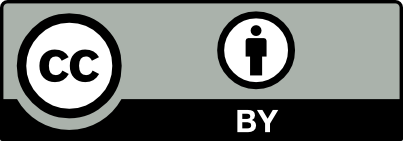}
\\ 
We gratefully acknowledge Polish high-performance
computing infrastructure PLGrid (HPC Center: ACK
Cyfronet AGH) for providing computer facilities and support
within computational grant no. \textbf{PLG/2025/018312}.

\bibliographystyle{IEEEtran}
\bibliography{custom}

\clearpage

\setcounter{page}{1}
\setcounter{section}{0}
\setcounter{equation}{0}
\setcounter{figure}{0}
\setcounter{table}{0}

\renewcommand{\theequation}{S\arabic{equation}}
\renewcommand{\thefigure}{S\arabic{figure}}
\renewcommand{\thetable}{S\arabic{table}}
\renewcommand{\thesection}{S\arabic{section}}

\twocolumn[{
\begin{center}
    \vspace*{2em}
    \Large \textit{Supplementary Material for:}\\[0.5em] 
    Bayesian Fine-tuning in Projected Subspaces \\[1.0em] 
    \large Viktar Dubovik, Patryk Marsza{\l}ek, Jacek Tabor, and Tomasz Ku{\'s}mierczyk \\[2em]
\end{center}
}]

\begin{abstract}
This supplement provides additional technical details and experimental results to support the findings of the main manuscript. 
Section~\ref{sec:experimental_details} details the Bayesian parameterization of our proposed variants (B-LoRA-XS, L-LoRA-XS, and L-LoRA-S), MAP training protocols, and implementation settings for RoBERTa-Large and LLaMA2-7B. 
Section~\ref{sec:additional_experiments} provides an extended ablation study on model robustness through a data size reduction analysis. 
Section~\ref{sec:numerical_results} contains comprehensive numerical results and performance metrics across multiple datasets, comparing various projection methods, adapter ranks, and covariance approximations.
\end{abstract}


\section{Experimental details}
\label{sec:experimental_details}

\subsection{Bayesian Parameterization}

Across all our Bayesian variants, posterior inference was performed solely on the square matrices $R$, and unlike traditional Bayesian LoRAs, 
the projection matrices $A$ and $B$ were not included for the posterior approximation.
Consequently, B-LoRA-XS, L-LoRA-XS, and L-LoRA-S differ in the posterior approximation method and/or MAP training procedure, while relying on the same set of random variables (e.g., posterior parameterization).

\subsection{Shared MAP Training Protocol}

Unless otherwise stated, MAP training followed the LoRA-XS setup~\cite{balazy2024lora}, including the AdamW optimizer~\cite{loshchilov2019}, a linear learning rate scheduler with warm-up, standard LoRA scaling, and the subset of target modules to which adapters were applied.
Differences between our methods arise from:
(i) the posterior approximation (SWAG or Laplace),
(ii) whether projection matrices $A$ and $B$ were fixed or optimized during MAP training, and
(iii) backbone-specific hyperparameters.

\subsection{B-LoRA-XS (SWAG variant)}

B-LoRA-XS follows the standard LoRA-XS MAP training procedure, with posterior uncertainty modeled using SWAG~\cite{NEURIPS2019_118921ef} over the square matrices $R$.

\paragraph{Posterior fitting}
After a burn-in phase corresponding to MAP training, we collect model checkpoints and fit a Gaussian approximation with a low-rank plus diagonal covariance. Unless otherwise stated, we used a covariance rank of $k=10$ and computed predictions by averaging over $S=15$ sampled models.

\newpage

\subsection{L-LoRA-XS (Laplace variant)}

L-LoRA-XS followed the standard LoRA-XS MAP training procedure and applied a post hoc Laplace approximation over the LoRA-XS square matrices $R$.

\paragraph{Curvature approximation}
We approximated the posterior with a Gaussian centered at the MAP solution (at an early stopped checkpoint), where curvature was given by Generalized Gauss-Newton (GGN) approximation. 
We considered both:
\begin{itemize}
\item KRON approximation with Curvlinops~\cite{dangel2025position} implementation,
\item DIAG approximation with ASDL~\cite{osawa2023asdlunifiedinterfacegradient} implementation.
\end{itemize}

\paragraph{Prior}
We assumed a Gaussian prior with scalar precision (inverse variance), corresponding to an $\ell_2$ regularizer in the MAP objective. The prior precision was optimized post hoc via marginal likelihood maximization.

\paragraph{Posterior fitting}
The Laplace approximation was applied post hoc at early stopping checkpoints.

\subsection{L-LoRA-S (Laplace with extended MAP optimization)}

L-LoRA-S differs from L-LoRA-XS in the MAP phase: the projection matrices $A$ and $B$ are additionally optimized during MAP training, which increases the expressive capacity of the adapter parameterization.
However, posterior inference remains identical to L-LoRA-XS and is restricted to the matrices $R$. Thus, L-LoRA-S modifies the MAP solution and consequently, the projected subspace, but not the posterior parametrization itself.
Therefore, \textbf{comparisons between L-LoRA-XS and L-LoRA-S isolates the effect of improving the MAP solution} via additional optimization of $A$ and $B$, while keeping the Bayesian parameterization fixed.

\paragraph{Configurations (LLaMA2-7B)}
For LLaMA2-7B, we considered two configurations:
\begin{itemize}
\item L-LoRA-S (10M): adapters were applied to \texttt{q\_proj}, \texttt{v\_proj}, \texttt{o\_proj}, and \texttt{down\_proj}.
\item L-LoRA-S (4M): adapters were restricted to \texttt{q\_proj} and \texttt{v\_proj}.
\end{itemize}

\subsection{RoBERTa-Large}
\label{sec:roberta}

For experiments involving RoBERTa-Large~\cite{liu2019}, we used pretrained checkpoints from the HuggingFace Transformers library~\cite{wolf2020}. For the RTE and MRPC datasets, we followed the LoRA-XS protocol~\cite{balazy2024lora}, initializing modules with weights fine-tuned on the MNLI task.

Adapters were inserted into the Query, Value, Attention Output, and Output Fully Connected weight matrices across all transformer layers~\cite{NIPS2017_3f5ee243}. Due to computational constraints, standard LoRA and LoRA-SWAG baselines were restricted to the Query and Value matrices only, which was sufficient to achieve their best performance.

\paragraph{SWAG details}
The SWAG starting epoch was set to 10 (SST-2 and MRPC) or 25 (other datasets).

\paragraph{Laplace details}
The Laplace approximation was applied at early stopping checkpoints:
\begin{itemize}
\item epoch 10 out of total 20 epochs (SST-2),
\item epoch 25 out of total 50 epochs (RTE, MRPC, CoLA).
\end{itemize}

\subsection{LLaMA2-7B}

To evaluate scalability, we studied LLaMA2-7B~\cite{touvron2023llama} with SVD-based projection.
In particular, we set:
\begin{itemize}
\item LoRA scaling factor $\alpha = 25$,
\item adapter dropout rate $= 0$,
\item weight decay $= 0.1$.
\end{itemize}

Learning rates for B-LoRA-XS and L-LoRA-XS were:
\begin{itemize}
\item $1 \times 10^{-3}$ for ARC-C and OBQA,
\item $5 \times 10^{-4}$ for ARC-E.
\end{itemize}
The learning rate for L-LoRA-S was $5 \times 10^{-4}$ on all datasets.

\paragraph{SWAG details}
SWAG collection started at epoch 10. The learning rate during SWAG was set to one tenth of the MAP learning rate at the selected checkpoint.

\paragraph{Laplace details}
MAP training was performed for 10 epochs. The Laplace approximation was fitted at the early stopping checkpoint (at the end of epoch 5).

\section{Additional Experiments}
\label{sec:additional_experiments}

\begin{figure}[t!]
    \centering
    {\scriptsize 
    \hspace{0.8cm} CoLA\\
    \vspace{0.025cm}
    \includegraphics[width=1.0\linewidth]{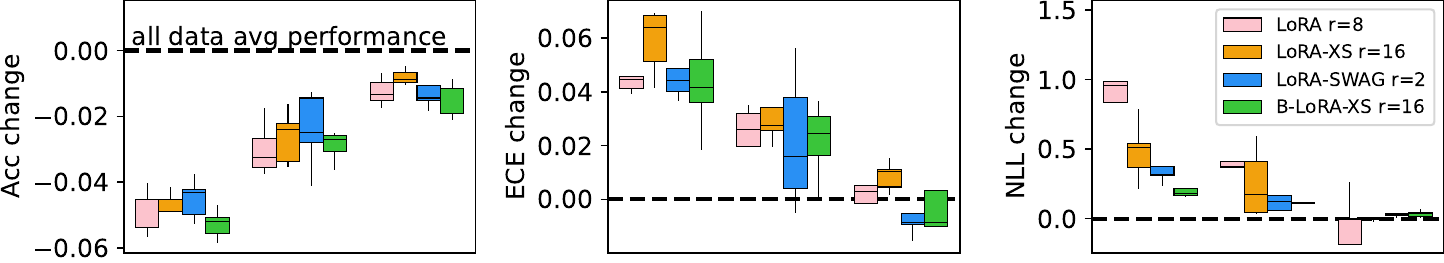} \\ 
    \vspace{0.01cm}
    \hspace{0.8cm} MRPC\\
    \vspace{0.025cm}
    \includegraphics[width=1.0\linewidth]{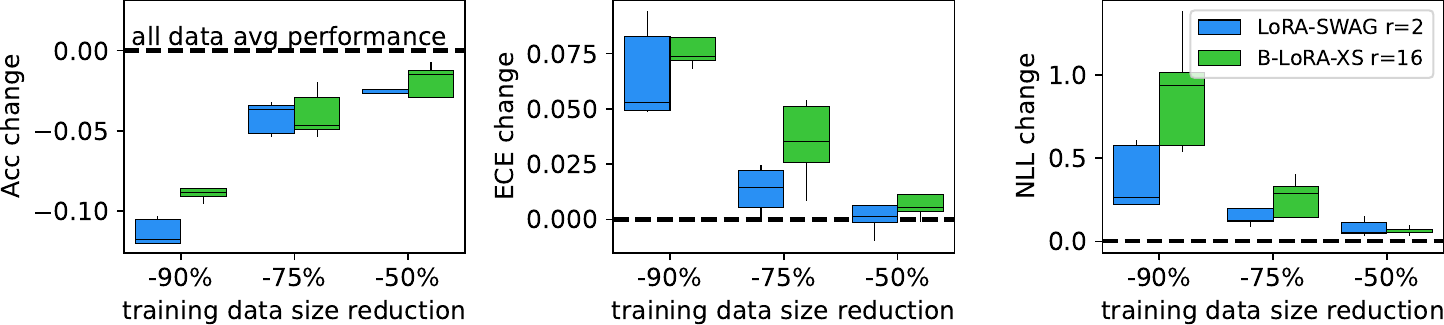}\\
    }
    \caption{Accuracy, ECE and NLL change as the training set is progressively reduced (e.g.\ -90\% means using only 10\% of the data for training). The dashed line marks the model's performance when trained on the full dataset.}
    \label{fig:subsampling}
\end{figure}

\subsection{Data Size Reduction Analysis}
\label{sec:data_reduction_analysis}
Fig.~\ref{fig:subsampling} compares how accuracy, ECE, and NLL degrade when training data is subsampled.  
All methods predictably lose accuracy as data size decreases, with little difference between the various LoRA-based approaches.  
We conclude that \textbf{Bayesian learning does not improve robustness in this case}.  
However, we note variations across datasets in terms of accuracy. For example, in MRPC, the decline is more pronounced, likely due to the dataset smaller size.  

\section{Numerical Results}
\label{sec:numerical_results}

\begin{table}
\caption{Laplace fine-tuning of RoBERTa with various  projections.
}
\label{tab:results_roberta_projections}
\centering
\scalebox{0.6}{
    \begin{tabular}{llll|lll}
    \toprule
    Dataset & rank $r$ & Covariance type & Projection & Accuracy $\uparrow$ & ECE $\downarrow$ & NLL $\downarrow$  \\
    \midrule
    \multirow[t]{28}{*}{CoLA} & \multirow[t]{14}{*}{8} & \multirow[t]{7}{*}{Diag} & W-SVD & 0.846 ${}_{0.018}$ & 0.045 ${}_{0.014}$ & 0.371 ${}_{0.030}$ \\
     &  &  & SVD & 0.846 ${}_{0.021}$ & 0.049 ${}_{0.017}$ & 0.378 ${}_{0.040}$ \\
     &  &  & DCT & 0.786 ${}_{0.022}$ & 0.106 ${}_{0.044}$ & 0.520 ${}_{0.083}$ \\
     &  &  & RAND & 0.706 ${}_{0.003}$ & 0.198 ${}_{0.020}$ & 0.692 ${}_{0.054}$ \\
     &  &  & WSVD-SVD & 0.691 ${}_{0.045}$ & 0.078 ${}_{0.036}$ & 0.619 ${}_{0.083}$ \\
     &  &  & DCT-SVD & 0.816 ${}_{0.001}$ & 0.070 ${}_{0.008}$ & 0.422 ${}_{0.008}$ \\
     &  &  & RAND-SVD & 0.802 ${}_{0.008}$ & 0.081 ${}_{0.031}$ & 0.427 ${}_{0.059}$ \\
    \cline{3-7}
     &  & \multirow[t]{7}{*}{Kron} & W-SVD & 0.846 ${}_{0.017}$ & 0.049 ${}_{0.011}$ & 0.376 ${}_{0.032}$ \\
     &  &  & SVD & 0.846 ${}_{0.021}$ & 0.054 ${}_{0.017}$ & 0.382 ${}_{0.042}$ \\
     &  &  & DCT & 0.786 ${}_{0.022}$ & 0.112 ${}_{0.041}$ & 0.533 ${}_{0.088}$ \\
     &  &  & RAND & 0.706 ${}_{0.003}$ & 0.199 ${}_{0.021}$ & 0.697 ${}_{0.058}$ \\
     &  &  & WSVD-SVD & 0.691 ${}_{0.044}$ & 0.080 ${}_{0.042}$ & 0.619 ${}_{0.073}$ \\
     &  &  & DCT-SVD & 0.816 ${}_{0.001}$ & 0.078 ${}_{0.010}$ & 0.427 ${}_{0.010}$ \\
     &  &  & RAND-SVD & 0.802 ${}_{0.008}$ & 0.088 ${}_{0.033}$ & 0.437 ${}_{0.063}$ \\
    \cline{2-7} \cline{3-7}
     & \multirow[t]{14}{*}{16} & \multirow[t]{7}{*}{Diag} & W-SVD & 0.833 ${}_{0.023}$ & 0.048 ${}_{0.020}$ & 0.377 ${}_{0.044}$ \\
     &  &  & SVD & 0.849 ${}_{0.008}$ & 0.029 ${}_{0.006}$ & 0.357 ${}_{0.004}$ \\
     &  &  & DCT & 0.815 ${}_{0.004}$ & 0.045 ${}_{0.010}$ & 0.420 ${}_{0.003}$ \\
     &  &  & RAND & 0.783 ${}_{0.024}$ & 0.055 ${}_{0.049}$ & 0.476 ${}_{0.048}$ \\
     &  &  & WSVD-SVD & 0.843 ${}_{0.011}$ & 0.034 ${}_{0.005}$ & 0.361 ${}_{0.014}$ \\
     &  &  & DCT-SVD & 0.839 ${}_{0.012}$ & 0.035 ${}_{0.007}$ & 0.368 ${}_{0.010}$ \\
     &  &  & RAND-SVD & 0.805 ${}_{0.016}$ & 0.059 ${}_{0.014}$ & 0.409 ${}_{0.016}$ \\
    \cline{3-7}
     &  & \multirow[t]{7}{*}{Kron} & W-SVD & 0.834 ${}_{0.023}$ & 0.051 ${}_{0.022}$ & 0.381 ${}_{0.048}$ \\
     &  &  & SVD & 0.849 ${}_{0.008}$ & 0.035 ${}_{0.007}$ & 0.359 ${}_{0.006}$ \\
     &  &  & DCT & 0.815 ${}_{0.004}$ & 0.046 ${}_{0.011}$ & 0.421 ${}_{0.001}$ \\
     &  &  & RAND & 0.782 ${}_{0.024}$ & 0.063 ${}_{0.048}$ & 0.478 ${}_{0.056}$ \\
     &  &  & WSVD-SVD & 0.843 ${}_{0.011}$ & 0.040 ${}_{0.006}$ & 0.369 ${}_{0.015}$ \\
     &  &  & DCT-SVD & 0.839 ${}_{0.012}$ & 0.039 ${}_{0.010}$ & 0.369 ${}_{0.011}$ \\
     &  &  & RAND-SVD & 0.805 ${}_{0.016}$ & 0.068 ${}_{0.015}$ & 0.416 ${}_{0.016}$ \\
    \cline{1-7} \cline{2-7} \cline{3-7}
    \multirow[t]{28}{*}{MRPC} & \multirow[t]{14}{*}{8} & \multirow[t]{7}{*}{Diag} & W-SVD & 0.858 ${}_{0.004}$ & 0.060 ${}_{0.008}$ & 0.327 ${}_{0.016}$ \\
     &  &  & SVD & 0.860 ${}_{0.010}$ & 0.051 ${}_{0.017}$ & 0.328 ${}_{0.012}$ \\
     &  &  & DCT & 0.860 ${}_{0.010}$ & 0.051 ${}_{0.017}$ & 0.328 ${}_{0.012}$ \\
     &  &  & RAND & 0.841 ${}_{0.019}$ & 0.041 ${}_{0.003}$ & 0.339 ${}_{0.001}$ \\
     &  &  & WSVD-SVD & 0.858 ${}_{0.004}$ & 0.060 ${}_{0.008}$ & 0.327 ${}_{0.016}$ \\
     &  &  & DCT-SVD & 0.860 ${}_{0.010}$ & 0.051 ${}_{0.017}$ & 0.328 ${}_{0.012}$ \\
     &  &  & RAND-SVD & 0.853 ${}_{0.005}$ & 0.044 ${}_{0.009}$ & 0.322 ${}_{0.010}$ \\
    \cline{3-7}
     &  & \multirow[t]{7}{*}{Kron} & W-SVD & 0.858 ${}_{0.005}$ & 0.048 ${}_{0.008}$ & 0.317 ${}_{0.016}$ \\
     &  &  & SVD & 0.858 ${}_{0.010}$ & 0.047 ${}_{0.017}$ & 0.324 ${}_{0.012}$ \\
     &  &  & DCT & 0.858 ${}_{0.010}$ & 0.047 ${}_{0.017}$ & 0.324 ${}_{0.012}$ \\
     &  &  & RAND & 0.841 ${}_{0.022}$ & 0.050 ${}_{0.009}$ & 0.338 ${}_{0.003}$ \\
     &  &  & WSVD-SVD & 0.858 ${}_{0.005}$ & 0.048 ${}_{0.008}$ & 0.317 ${}_{0.016}$ \\
     &  &  & DCT-SVD & 0.858 ${}_{0.010}$ & 0.047 ${}_{0.017}$ & 0.324 ${}_{0.012}$ \\
     &  &  & RAND-SVD & 0.855 ${}_{0.004}$ & 0.047 ${}_{0.021}$ & 0.320 ${}_{0.008}$ \\
    \cline{2-7} \cline{3-7}
     & \multirow[t]{14}{*}{16} & \multirow[t]{7}{*}{Diag} & W-SVD & 0.887 ${}_{0.011}$ & 0.063 ${}_{0.006}$ & 0.286 ${}_{0.017}$ \\
     &  &  & SVD & 0.895 ${}_{0.006}$ & 0.077 ${}_{0.018}$ & 0.293 ${}_{0.011}$ \\
     &  &  & DCT & 0.895 ${}_{0.006}$ & 0.077 ${}_{0.018}$ & 0.293 ${}_{0.011}$ \\
     &  &  & RAND & 0.890 ${}_{0.010}$ & 0.085 ${}_{0.018}$ & 0.300 ${}_{0.026}$ \\
     &  &  & WSVD-SVD & 0.887 ${}_{0.011}$ & 0.063 ${}_{0.006}$ & 0.286 ${}_{0.017}$ \\
     &  &  & DCT-SVD & 0.895 ${}_{0.006}$ & 0.077 ${}_{0.018}$ & 0.293 ${}_{0.011}$ \\
     &  &  & RAND-SVD & 0.885 ${}_{0.005}$ & 0.096 ${}_{0.037}$ & 0.304 ${}_{0.031}$ \\
    \cline{3-7}
     &  & \multirow[t]{7}{*}{Kron} & W-SVD & 0.887 ${}_{0.011}$ & 0.051 ${}_{0.006}$ & 0.279 ${}_{0.015}$ \\
     &  &  & SVD & 0.892 ${}_{0.008}$ & 0.066 ${}_{0.021}$ & 0.286 ${}_{0.011}$ \\
     &  &  & DCT & 0.892 ${}_{0.008}$ & 0.066 ${}_{0.021}$ & 0.286 ${}_{0.011}$ \\
     &  &  & RAND & 0.890 ${}_{0.007}$ & 0.069 ${}_{0.018}$ & 0.287 ${}_{0.024}$ \\
     &  &  & WSVD-SVD & 0.887 ${}_{0.011}$ & 0.051 ${}_{0.006}$ & 0.279 ${}_{0.015}$ \\
     &  &  & DCT-SVD & 0.892 ${}_{0.008}$ & 0.066 ${}_{0.021}$ & 0.286 ${}_{0.011}$ \\
     &  &  & RAND-SVD & 0.887 ${}_{0.004}$ & 0.084 ${}_{0.028}$ & 0.292 ${}_{0.027}$ \\
    \cline{1-7} \cline{2-7} \cline{3-7}
    \multirow[t]{28}{*}{RTE} & \multirow[t]{14}{*}{8} & \multirow[t]{7}{*}{Diag} & W-SVD & 0.841 ${}_{0.006}$ & 0.096 ${}_{0.018}$ & 0.385 ${}_{0.007}$ \\
     &  &  & SVD & 0.856 ${}_{0.008}$ & 0.099 ${}_{0.019}$ & 0.383 ${}_{0.009}$ \\
     &  &  & DCT & 0.856 ${}_{0.018}$ & 0.100 ${}_{0.014}$ & 0.378 ${}_{0.005}$ \\
     &  &  & RAND & 0.841 ${}_{0.000}$ & 0.107 ${}_{0.009}$ & 0.390 ${}_{0.005}$ \\
     &  &  & WSVD-SVD & 0.852 ${}_{0.004}$ & 0.107 ${}_{0.013}$ & 0.385 ${}_{0.010}$ \\
     &  &  & DCT-SVD & 0.845 ${}_{0.014}$ & 0.090 ${}_{0.013}$ & 0.382 ${}_{0.005}$ \\
     &  &  & RAND-SVD & 0.859 ${}_{0.028}$ & 0.104 ${}_{0.009}$ & 0.381 ${}_{0.005}$ \\
    \cline{3-7}
     &  & \multirow[t]{7}{*}{Kron} & W-SVD & 0.838 ${}_{0.006}$ & 0.090 ${}_{0.007}$ & 0.381 ${}_{0.008}$ \\
     &  &  & SVD & 0.852 ${}_{0.011}$ & 0.088 ${}_{0.021}$ & 0.378 ${}_{0.009}$ \\
     &  &  & DCT & 0.856 ${}_{0.017}$ & 0.094 ${}_{0.013}$ & 0.375 ${}_{0.005}$ \\
     &  &  & RAND & 0.841 ${}_{0.000}$ & 0.100 ${}_{0.012}$ & 0.386 ${}_{0.005}$ \\
     &  &  & WSVD-SVD & 0.852 ${}_{0.006}$ & 0.096 ${}_{0.013}$ & 0.381 ${}_{0.010}$ \\
     &  &  & DCT-SVD & 0.845 ${}_{0.016}$ & 0.087 ${}_{0.013}$ & 0.382 ${}_{0.005}$ \\
     &  &  & RAND-SVD & 0.856 ${}_{0.026}$ & 0.100 ${}_{0.003}$ & 0.377 ${}_{0.007}$ \\
    \cline{2-7} \cline{3-7}
     & \multirow[t]{14}{*}{16} & \multirow[t]{7}{*}{Diag} & W-SVD & 0.874 ${}_{0.004}$ & 0.106 ${}_{0.002}$ & 0.352 ${}_{0.003}$ \\
     &  &  & SVD & 0.874 ${}_{0.011}$ & 0.112 ${}_{0.007}$ & 0.355 ${}_{0.012}$ \\
     &  &  & DCT & 0.874 ${}_{0.009}$ & 0.116 ${}_{0.005}$ & 0.364 ${}_{0.007}$ \\
     &  &  & RAND & 0.881 ${}_{0.010}$ & 0.111 ${}_{0.006}$ & 0.350 ${}_{0.003}$ \\
     &  &  & WSVD-SVD & 0.874 ${}_{0.002}$ & 0.109 ${}_{0.009}$ & 0.354 ${}_{0.012}$ \\
     &  &  & DCT-SVD & 0.874 ${}_{0.010}$ & 0.113 ${}_{0.008}$ & 0.359 ${}_{0.004}$ \\
     &  &  & RAND-SVD & 0.866 ${}_{0.010}$ & 0.118 ${}_{0.017}$ & 0.356 ${}_{0.006}$ \\
    \cline{3-7}
     &  & \multirow[t]{7}{*}{Kron} & W-SVD & 0.870 ${}_{0.004}$ & 0.098 ${}_{0.004}$ & 0.345 ${}_{0.003}$ \\
     &  &  & SVD & 0.874 ${}_{0.008}$ & 0.103 ${}_{0.006}$ & 0.353 ${}_{0.010}$ \\
     &  &  & DCT & 0.874 ${}_{0.011}$ & 0.109 ${}_{0.005}$ & 0.358 ${}_{0.006}$ \\
     &  &  & RAND & 0.877 ${}_{0.009}$ & 0.108 ${}_{0.002}$ & 0.348 ${}_{0.002}$ \\
     &  &  & WSVD-SVD & 0.877 ${}_{0.002}$ & 0.105 ${}_{0.008}$ & 0.350 ${}_{0.008}$ \\
     &  &  & DCT-SVD & 0.874 ${}_{0.010}$ & 0.106 ${}_{0.009}$ & 0.357 ${}_{0.003}$ \\
     &  &  & RAND-SVD & 0.866 ${}_{0.010}$ & 0.109 ${}_{0.017}$ & 0.353 ${}_{0.005}$ \\
    \cline{1-7} \cline{2-7} \cline{3-7}
    \multirow[t]{28}{*}{SST-2} & \multirow[t]{14}{*}{8} & \multirow[t]{7}{*}{Diag} & W-SVD & 0.953 ${}_{0.001}$ & 0.018 ${}_{0.006}$ & 0.138 ${}_{0.003}$ \\
     &  &  & SVD & 0.945 ${}_{0.005}$ & 0.017 ${}_{0.003}$ & 0.154 ${}_{0.001}$ \\
     &  &  & DCT & 0.940 ${}_{0.001}$ & 0.021 ${}_{0.005}$ & 0.168 ${}_{0.002}$ \\
     &  &  & RAND & 0.930 ${}_{0.002}$ & 0.019 ${}_{0.002}$ & 0.188 ${}_{0.007}$ \\
     &  &  & WSVD-SVD & 0.727 ${}_{0.308}$ & 0.048 ${}_{0.048}$ & 0.428 ${}_{0.394}$ \\
     &  &  & DCT-SVD & 0.943 ${}_{0.002}$ & 0.015 ${}_{0.003}$ & 0.150 ${}_{0.004}$ \\
     &  &  & RAND-SVD & 0.945 ${}_{0.005}$ & 0.019 ${}_{0.001}$ & 0.152 ${}_{0.003}$ \\
    \cline{3-7}
     &  & \multirow[t]{7}{*}{Kron} & W-SVD & 0.953 ${}_{0.001}$ & 0.021 ${}_{0.005}$ & 0.139 ${}_{0.002}$ \\
     &  &  & SVD & 0.945 ${}_{0.005}$ & 0.017 ${}_{0.005}$ & 0.154 ${}_{0.001}$ \\
     &  &  & DCT & 0.940 ${}_{0.001}$ & 0.020 ${}_{0.004}$ & 0.169 ${}_{0.003}$ \\
     &  &  & RAND & 0.930 ${}_{0.002}$ & 0.016 ${}_{0.005}$ & 0.188 ${}_{0.007}$ \\
     &  &  & WSVD-SVD & 0.727 ${}_{0.308}$ & 0.051 ${}_{0.045}$ & 0.429 ${}_{0.394}$ \\
     &  &  & DCT-SVD & 0.943 ${}_{0.002}$ & 0.015 ${}_{0.002}$ & 0.150 ${}_{0.004}$ \\
     &  &  & RAND-SVD & 0.945 ${}_{0.005}$ & 0.019 ${}_{0.001}$ & 0.152 ${}_{0.003}$ \\
    \cline{2-7} \cline{3-7}
     & \multirow[t]{14}{*}{16} & \multirow[t]{7}{*}{Diag} & W-SVD & 0.954 ${}_{0.005}$ & 0.021 ${}_{0.002}$ & 0.138 ${}_{0.006}$ \\
     &  &  & SVD & 0.953 ${}_{0.001}$ & 0.017 ${}_{0.005}$ & 0.137 ${}_{0.000}$ \\
     &  &  & DCT & 0.950 ${}_{0.003}$ & 0.018 ${}_{0.002}$ & 0.144 ${}_{0.003}$ \\
     &  &  & RAND & 0.939 ${}_{0.002}$ & 0.014 ${}_{0.005}$ & 0.162 ${}_{0.004}$ \\
     &  &  & WSVD-SVD & 0.958 ${}_{0.004}$ & 0.015 ${}_{0.004}$ & 0.133 ${}_{0.004}$ \\
     &  &  & DCT-SVD & 0.953 ${}_{0.003}$ & 0.021 ${}_{0.004}$ & 0.139 ${}_{0.001}$ \\
     &  &  & RAND-SVD & 0.953 ${}_{0.004}$ & 0.015 ${}_{0.001}$ & 0.141 ${}_{0.002}$ \\
    \cline{3-7}
     &  & \multirow[t]{7}{*}{Kron} & W-SVD & 0.954 ${}_{0.005}$ & 0.021 ${}_{0.005}$ & 0.139 ${}_{0.007}$ \\
     &  &  & SVD & 0.953 ${}_{0.001}$ & 0.015 ${}_{0.002}$ & 0.136 ${}_{0.000}$ \\
     &  &  & DCT & 0.950 ${}_{0.002}$ & 0.017 ${}_{0.001}$ & 0.145 ${}_{0.003}$ \\
     &  &  & RAND & 0.939 ${}_{0.002}$ & 0.016 ${}_{0.004}$ & 0.162 ${}_{0.004}$ \\
     &  &  & WSVD-SVD & 0.958 ${}_{0.004}$ & 0.016 ${}_{0.005}$ & 0.133 ${}_{0.003}$ \\
     &  &  & DCT-SVD & 0.953 ${}_{0.004}$ & 0.015 ${}_{0.003}$ & 0.138 ${}_{0.001}$ \\
     &  &  & RAND-SVD & 0.953 ${}_{0.003}$ & 0.019 ${}_{0.004}$ & 0.141 ${}_{0.002}$ \\
    \bottomrule
    \end{tabular}
}
\end{table}

\begin{table}
\caption{Laplace fine-tuning of LLaMA2-7B with various  projections.
}
\label{tab:results_llama_projections}
\centering
\resizebox{1.0\linewidth}{!}{
\begin{tabular}{llll|lll}
\toprule
Dataset & rank $r$ & Covariance type & Projection & Accuracy $\uparrow$ & ECE $\downarrow$ & NLL $\downarrow$ \\
\midrule
\multirow[t]{16}{*}{ARC-E} & \multirow[t]{8}{*}{32} & \multirow[t]{4}{*}{DIAG} & W-SVD & 0.838 ${}_{0.011}$ & 0.099 ${}_{0.007}$ & 0.496 ${}_{0.014}$ \\
 &  &  & SVD & 0.818 ${}_{0.010}$ & 0.086 ${}_{0.007}$ & 0.535 ${}_{0.006}$ \\
 &  &  & DCT & 0.388 ${}_{0.006}$ & 0.131 ${}_{0.008}$ & 1.287 ${}_{0.004}$ \\
 &  &  & RAND & 0.379 ${}_{0.004}$ & 0.162 ${}_{0.004}$ & 1.347 ${}_{0.005}$ \\
\cline{3-7}
 &  & \multirow[t]{4}{*}{KRON} & W-SVD & 0.838 ${}_{0.012}$ & 0.067 ${}_{0.006}$ & 0.479 ${}_{0.016}$ \\
 &  &  & SVD & 0.818 ${}_{0.008}$ & 0.053 ${}_{0.005}$ & 0.527 ${}_{0.006}$ \\
 &  &  & DCT & 0.390 ${}_{0.006}$ & 0.136 ${}_{0.010}$ & 1.292 ${}_{0.003}$ \\
 &  &  & RAND & 0.379 ${}_{0.003}$ & 0.171 ${}_{0.005}$ & 1.353 ${}_{0.005}$ \\
\cline{2-7} \cline{3-7}
 & \multirow[t]{8}{*}{64} & \multirow[t]{4}{*}{DIAG} & W-SVD & 0.836 ${}_{0.009}$ & 0.130 ${}_{0.005}$ & 0.546 ${}_{0.014}$ \\
 &  &  & SVD & 0.820 ${}_{0.005}$ & 0.133 ${}_{0.002}$ & 0.572 ${}_{0.009}$ \\
 &  &  & DCT & 0.407 ${}_{0.004}$ & 0.074 ${}_{0.009}$ & 1.242 ${}_{0.003}$ \\
 &  &  & RAND & 0.406 ${}_{0.003}$ & 0.117 ${}_{0.005}$ & 1.275 ${}_{0.007}$ \\
\cline{3-7}
 &  & \multirow[t]{4}{*}{KRON} & W-SVD & 0.834 ${}_{0.010}$ & 0.089 ${}_{0.005}$ & 0.514 ${}_{0.019}$ \\
 &  &  & SVD & 0.820 ${}_{0.004}$ & 0.074 ${}_{0.006}$ & 0.531 ${}_{0.010}$ \\
 &  &  & DCT & 0.407 ${}_{0.006}$ & 0.077 ${}_{0.009}$ & 1.245 ${}_{0.002}$ \\
 &  &  & RAND & 0.404 ${}_{0.004}$ & 0.135 ${}_{0.007}$ & 1.282 ${}_{0.008}$ \\
\cline{1-7} \cline{2-7} \cline{3-7}
\multirow[t]{16}{*}{OBQA} & \multirow[t]{8}{*}{32} & \multirow[t]{4}{*}{DIAG} & W-SVD & 0.762 ${}_{0.012}$ & 0.063 ${}_{0.010}$ & 0.664 ${}_{0.027}$ \\
 &  &  & SVD & 0.766 ${}_{0.296}$ & 0.062 ${}_{0.013}$ & 0.693 ${}_{0.420}$ \\
 &  &  & DCT & 0.466 ${}_{0.006}$ & 0.074 ${}_{0.004}$ & 1.199 ${}_{0.001}$ \\
 &  &  & RAND & 0.434 ${}_{0.005}$ & 0.076 ${}_{0.006}$ & 1.239 ${}_{0.002}$ \\
\cline{3-7}
 &  & \multirow[t]{4}{*}{KRON} & W-SVD & 0.762 ${}_{0.009}$ & 0.042 ${}_{0.006}$ & 0.671 ${}_{0.032}$ \\
 &  &  & SVD & 0.766 ${}_{0.297}$ & 0.070 ${}_{0.001}$ & 0.701 ${}_{0.418}$ \\
 &  &  & DCT & 0.468 ${}_{0.007}$ & 0.064 ${}_{0.005}$ & 1.208 ${}_{0.000}$ \\
 &  &  & RAND & 0.436 ${}_{0.008}$ & 0.087 ${}_{0.001}$ & 1.247 ${}_{0.002}$ \\
\cline{2-7} \cline{3-7}
 & \multirow[t]{8}{*}{64} & \multirow[t]{4}{*}{DIAG} & W-SVD & 0.790 ${}_{0.003}$ & 0.068 ${}_{0.010}$ & 0.642 ${}_{0.014}$ \\
 &  &  & SVD & 0.774 ${}_{0.005}$ & 0.092 ${}_{0.012}$ & 0.652 ${}_{0.007}$ \\
 &  &  & DCT & 0.604 ${}_{0.001}$ & 0.058 ${}_{0.008}$ & 0.951 ${}_{0.003}$ \\
 &  &  & RAND & 0.546 ${}_{0.021}$ & 0.104 ${}_{0.011}$ & 1.145 ${}_{0.013}$ \\
\cline{3-7}
 &  & \multirow[t]{4}{*}{KRON} & W-SVD & 0.790 ${}_{0.003}$ & 0.061 ${}_{0.007}$ & 0.643 ${}_{0.017}$ \\
 &  &  & SVD & 0.772 ${}_{0.003}$ & 0.064 ${}_{0.002}$ & 0.652 ${}_{0.008}$ \\
 &  &  & DCT & 0.604 ${}_{0.004}$ & 0.066 ${}_{0.009}$ & 0.959 ${}_{0.004}$ \\
 &  &  & RAND & 0.542 ${}_{0.020}$ & 0.120 ${}_{0.015}$ & 1.178 ${}_{0.009}$ \\
\bottomrule
\end{tabular}
}
\end{table}

\begin{table}
\centering
\caption{Numeric values for  RoBERTa-Large (with SVD) on CoLA data}
\label{tab:num_cola}
{
\begin{tabular}{lll|rrr}
\toprule
method & $r$ & $k$ & Accuracy $\uparrow$ & ECE $\downarrow$ & NLL $\downarrow$ \\
\midrule
LoRA & 2 & 10 & $0.865_{0.004}$ & $0.116_{0.044}$ & $0.652_{0.215}$ \\
 & 8 & 10 & $0.870_{0.004}$ & $0.125_{0.008}$ & $0.933_{0.154}$ \\
LoRA-SWAG & 2 & 2 & $0.870_{0.003}$ & $0.086_{0.013}$ & $0.390_{0.018}$ \\
 & 2 & 10 & $0.870_{0.002}$ & $0.049_{0.013}$ & $0.365_{0.015}$ \\
 & 8 & 2 & $0.875_{0.005}$ & $0.078_{0.016}$ & $0.384_{0.012}$ \\
 & 8 & 10 & $0.874_{0.004}$ & $0.037_{0.008}$ & $0.351_{0.009}$ \\
LoRA-XS & 2 & 10 & $0.822_{0.009}$ & $0.065_{0.023}$ & $0.451_{0.022}$ \\
 & 8 & 10 & $0.853_{0.002}$ & $0.059_{0.021}$ & $0.440_{0.052}$ \\
 & 16 & 10 & $0.859_{0.003}$ & $0.096_{0.016}$ & $0.516_{0.067}$ \\
 & 25 & 10 & $0.869_{0.002}$ & $0.099_{0.021}$ & $0.465_{0.102}$ \\
B-LoRA-XS & 2 & 5 & $0.822_{0.002}$ & $0.040_{0.009}$ & $0.412_{0.003}$ \\
 & 2 & 10 & $0.822_{0.005}$ & $0.036_{0.017}$ & $0.422_{0.016}$ \\
 & 8 & 10 & $0.855_{0.004}$ & $0.044_{0.005}$ & $0.372_{0.018}$ \\
 & 16 & 5 & $0.863_{0.003}$ & $0.038_{0.020}$ & $0.354_{0.037}$ \\
 & 16 & 10 & $0.863_{0.001}$ & $0.046_{0.007}$ & $0.367_{0.006}$ \\
 & 25 & 5 & $0.870_{0.003}$ & $0.041_{0.006}$ & $0.360_{0.021}$ \\
 & 25 & 10 & $0.869_{0.002}$ & $0.049_{0.013}$ & $0.378_{0.016}$ \\
L-LoRA-XS & 2 & KRON & $0.809_{0.006}$ & $0.034_{0.007}$ & $0.429_{0.008}$ \\
 & 8 & KRON & $0.849_{0.003}$ & $0.030_{0.007}$ & $0.359_{0.011}$ \\
 & 16 & KRON & $0.851_{0.009}$ & $0.040_{0.008}$ & $0.373_{0.009}$ \\
 & 25 & KRON & $0.863_{0.005}$ & $0.038_{0.015}$ & $0.356_{0.015}$ \\
\bottomrule
\end{tabular}
}
\end{table}

\begin{table}
\centering
\caption{Numeric values for  RoBERTa-Large (with SVD) on  MPRC data}
\label{tab:num_mrpc}
{
\begin{tabular}{lll|rrr}
\toprule
method & $r$ & $k$ & Accuracy $\uparrow$ & ECE $\downarrow$ & NLL $\downarrow$ \\
\midrule
LoRA & 2 & 10 & $0.912_{0.003}$ & $0.069_{0.010}$ & $0.406_{0.230}$ \\
 & 8 & 10 & $0.912_{0.005}$ & $0.086_{0.006}$ & $0.727_{0.165}$ \\
LoRA-SWAG & 2 & 2 & $0.917_{0.004}$ & $0.112_{0.035}$ & $0.332_{0.034}$ \\
 & 2 & 10 & $0.917_{0.005}$ & $0.052_{0.016}$ & $0.306_{0.031}$ \\
 & 8 & 2 & $0.912_{0.003}$ & $0.056_{0.035}$ & $0.331_{0.031}$ \\
 & 8 & 10 & $0.912_{0.004}$ & $0.048_{0.018}$ & $0.321_{0.127}$ \\
LoRA-XS & 2 & 10 & $0.861_{0.017}$ & $0.048_{0.010}$ & $0.338_{0.022}$ \\
 & 8 & 10 & $0.886_{0.007}$ & $0.078_{0.023}$ & $0.355_{0.105}$ \\
 & 16 & 10 & $0.904_{0.006}$ & $0.079_{0.015}$ & $0.450_{0.135}$ \\
 & 25 & 10 & $0.904_{0.008}$ & $0.088_{0.015}$ & $0.560_{0.176}$ \\
B-LoRA-XS & 2 & 2 & $0.860_{0.012}$ & $0.080_{0.029}$ & $0.386_{0.027}$ \\
 & 2 & 10 & $0.858_{0.012}$ & $0.046_{0.011}$ & $0.336_{0.025}$ \\
 & 8 & 10 & $0.890_{0.004}$ & $0.043_{0.014}$ & $0.304_{0.023}$ \\
 & 16 & 2 & $0.912_{0.003}$ & $0.037_{0.010}$ & $0.270_{0.030}$ \\
 & 16 & 10 & $0.909_{0.007}$ & $0.047_{0.007}$ & $0.301_{0.044}$ \\
 & 25 & 2 & $0.917_{0.005}$ & $0.036_{0.011}$ & $0.268_{0.020}$ \\
 & 25 & 10 & $0.909_{0.005}$ & $0.049_{0.004}$ & $0.312_{0.013}$ \\
L-LoRA-XS & 2 & KRON & $0.855_{0.019}$ & $0.036_{0.007}$ & $0.344_{0.023}$ \\
 & 8 & KRON & $0.875_{0.017}$ & $0.025_{0.008}$ & $0.308_{0.015}$ \\
 & 16 & KRON & $0.907_{0.008}$ & $0.047_{0.018}$ & $0.263_{0.010}$ \\
 & 25 & KRON & $0.909_{0.001}$ & $0.046_{0.016}$ & $0.261_{0.014}$ \\
\bottomrule
\end{tabular}
}
\end{table}

\begin{table}
\centering
\caption{Numeric values for  RoBERTa-Large (with SVD) on  RTE data}
\label{tab:num_rte}
{
\begin{tabular}{lll|rrr}
\toprule
method & $r$ & $k$ & Accuracy $\uparrow$ & ECE $\downarrow$ & NLL $\downarrow$ \\
\midrule
LoRA & 2 & 10 & $0.874_{0.008}$ & $0.123_{0.007}$ & $1.264_{0.239}$ \\
 & 8 & 10 & $0.874_{0.010}$ & $0.125_{0.010}$ & $1.072_{0.123}$ \\
LoRA-SWAG & 2 & 10 & $0.870_{0.007}$ & $0.091_{0.009}$ & $0.518_{0.046}$ \\
 & 8 & 10 & $0.877_{0.011}$ & $0.078_{0.009}$ & $0.388_{0.039}$ \\
LoRA-XS & 2 & 10 & $0.632_{0.043}$ & $0.126_{0.046}$ & $0.730_{0.052}$ \\
 & 8 & 10 & $0.870_{0.005}$ & $0.116_{0.017}$ & $0.698_{0.188}$ \\
 & 16 & 10 & $0.884_{0.007}$ & $0.097_{0.013}$ & $0.644_{0.123}$ \\
 & 25 & 10 & $0.902_{0.005}$ & $0.099_{0.006}$ & $0.957_{0.045}$ \\
B-LoRA-XS & 2 & 10 & $0.650_{0.062}$ & $0.079_{0.025}$ & $0.652_{0.024}$ \\
 & 8 & 10 & $0.877_{0.003}$ & $0.083_{0.004}$ & $0.465_{0.029}$ \\
 & 16 & 10 & $0.888_{0.007}$ & $0.073_{0.011}$ & $0.446_{0.030}$ \\
 & 25 & 10 & $0.892_{0.005}$ & $0.076_{0.022}$ & $0.510_{0.239}$ \\
L-LoRA-XS & 2 & KRON & $0.621_{0.063}$ & $0.061_{0.030}$ & $0.655_{0.040}$ \\
 & 8 & KRON & $0.845_{0.012}$ & $0.046_{0.012}$ & $0.371_{0.012}$ \\
 & 16 & KRON & $0.888_{0.006}$ & $0.059_{0.011}$ & $0.304_{0.017}$ \\
 & 25 & KRON & $0.884_{0.009}$ & $0.073_{0.018}$ & $0.321_{0.013}$ \\
\bottomrule
\end{tabular}
}
\end{table}

\begin{table}
\centering
\caption{Numeric values for RoBERTa-Large (with SVD) on  SST-2 data}
\label{tab:num_sst2}
{
\begin{tabular}{lll|rrr}
\toprule
method & $r$ & $k$ & Accuracy $\uparrow$ & ECE $\downarrow$ & NLL $\downarrow$ \\
\midrule
LoRA & 2 & 10 & $0.961_{0.003}$ & $0.030_{0.006}$ & $0.162_{0.027}$ \\
 & 8 & 10 & $0.962_{0.002}$ & $0.032_{0.004}$ & $0.198_{0.025}$ \\
LoRA-SWAG & 2 & 10 & $0.956_{0.003}$ & $0.020_{0.069}$ & $0.145_{0.068}$ \\
 & 8 & 10 & $0.966_{0.004}$ & $0.030_{0.033}$ & $0.141_{0.031}$ \\
LoRA-XS & 2 & 10 & $0.944_{0.001}$ & $0.026_{0.002}$ & $0.168_{0.003}$ \\
 & 8 & 10 & $0.953_{0.002}$ & $0.034_{0.005}$ & $0.175_{0.011}$ \\
 & 16 & 10 & $0.959_{0.002}$ & $0.032_{0.003}$ & $0.161_{0.012}$ \\
 & 25 & 10 & $0.958_{0.003}$ & $0.032_{0.005}$ & $0.160_{0.021}$ \\
B-LoRA-XS & 2 & 10 & $0.945_{0.002}$ & $0.025_{0.003}$ & $0.163_{0.003}$ \\
 & 8 & 10 & $0.952_{0.001}$ & $0.019_{0.006}$ & $0.152_{0.008}$ \\
 & 16 & 10 & $0.958_{0.001}$ & $0.025_{0.006}$ & $0.147_{0.014}$ \\
 & 25 & 10 & $0.961_{0.000}$ & $0.027_{0.005}$ & $0.137_{0.007}$ \\
L-LoRA-XS & 8 & KRON & $0.946_{0.001}$ & $0.018_{0.003}$ & $0.149_{0.006}$ \\
 & 16 & KRON & $0.956_{0.003}$ & $0.014_{0.002}$ & $0.131_{0.001}$ \\
\bottomrule
\end{tabular}
}
\end{table}

\begin{table}
\centering 
\caption{Covariance matrix rank $k$ analysis for CoLA data}
\label{tab:cov_cola}
{
\begin{tabular}{lll|rrr}
\toprule
method & $r$ & $k$ & Accuracy $\uparrow$ & ECE $\downarrow$ & NLL $\downarrow$ \\
\midrule
LoRA & 2 & - & $0.865_{0.004}$ & $0.116_{0.047}$ & $0.652_{0.228}$ \\
LoRA-SWAG & 2 & 0 & $0.859_{0.002}$ & $0.117_{0.007}$ & $0.675_{0.067}$ \\
 & 2 & 2 & $0.870_{0.003}$ & $0.086_{0.013}$ & $0.390_{0.018}$ \\
 & 2 & 5 & $0.869_{0.002}$ & $0.047_{0.012}$ & $0.362_{0.010}$ \\
 & 2 & 10 & $0.870_{0.003}$ & $0.049_{0.014}$ & $0.365_{0.016}$ \\
 & 2 & 20 & $0.870_{0.005}$ & $0.039_{0.003}$ & $0.373_{0.018}$ \\
 & 8 & 0 & $0.864_{0.006}$ & $0.122_{0.008}$ & $0.773_{0.064}$ \\
 & 8 & 2 & $0.875_{0.005}$ & $0.078_{0.016}$ & $0.384_{0.012}$ \\
 & 8 & 5 & $0.873_{0.005}$ & $0.051_{0.008}$ & $0.364_{0.013}$ \\
 & 8 & 10 & $0.874_{0.004}$ & $0.037_{0.009}$ & $0.351_{0.010}$ \\
 & 8 & 20 & $0.876_{0.005}$ & $0.032_{0.019}$ & $0.360_{0.034}$ \\
LoRA-XS & 16 & - & $0.859_{0.004}$ & $0.096_{0.017}$ & $0.516_{0.071}$ \\
B-LoRA-XS & 16 & 0 & $0.865_{0.004}$ & $0.068_{0.008}$ & $0.396_{0.021}$ \\
 & 16 & 2 & $0.864_{0.007}$ & $0.047_{0.016}$ & $0.372_{0.031}$ \\
 & 16 & 5 & $0.863_{0.003}$ & $0.038_{0.020}$ & $0.354_{0.037}$ \\
 & 16 & 10 & $0.863_{0.001}$ & $0.046_{0.008}$ & $0.367_{0.006}$ \\
 & 16 & 20 & $0.861_{0.003}$ & $0.051_{0.005}$ & $0.360_{0.015}$ \\
\bottomrule
\end{tabular}
}
\end{table}

\begin{table}
\centering
\caption{Covariance matrix rank $k$ analysis for MRPC data}
\label{tab:cov_mrpc}
{
\begin{tabular}{lll|rrr}
\toprule
method & $r$ & $k$ & Accuracy $\uparrow$ & ECE $\downarrow$ & NLL $\downarrow$ \\
\midrule
LoRA & 2 & - & $0.912_{0.003}$ & $0.069_{0.011}$ & $0.406_{0.244}$ \\
LoRA-SWAG & 2 & 0 & $0.904_{0.003}$ & $0.089_{0.003}$ & $0.628_{0.077}$ \\
 & 2 & 2 & $0.914_{0.004}$ & $0.135_{0.045}$ & $0.340_{0.040}$ \\
 & 2 & 5 & $0.914_{0.007}$ & $0.048_{0.005}$ & $0.294_{0.028}$ \\
 & 2 & 10 & $0.917_{0.005}$ & $0.052_{0.017}$ & $0.306_{0.033}$ \\
 & 2 & 20 & $0.914_{0.005}$ & $0.051_{0.015}$ & $0.306_{0.086}$ \\
 & 8 & 0 & $0.907_{0.005}$ & $0.094_{0.005}$ & $0.759_{0.123}$ \\
 & 8 & 2 & $0.912_{0.003}$ & $0.056_{0.035}$ & $0.331_{0.031}$ \\
 & 8 & 5 & $0.914_{0.003}$ & $0.040_{0.019}$ & $0.283_{0.103}$ \\
 & 8 & 10 & $0.917_{0.004}$ & $0.051_{0.017}$ & $0.328_{0.128}$ \\
 & 8 & 20 & $0.909_{0.006}$ & $0.053_{0.018}$ & $0.314_{0.177}$ \\
LoRA-XS & 16 & - & $0.909_{0.007}$ & $0.079_{0.011}$ & $0.388_{0.095}$ \\
B-LoRA-XS & 16 & 0 & $0.907_{0.006}$ & $0.078_{0.009}$ & $0.426_{0.037}$ \\
 & 16 & 2 & $0.912_{0.003}$ & $0.037_{0.010}$ & $0.270_{0.030}$ \\
 & 16 & 5 & $0.909_{0.009}$ & $0.051_{0.010}$ & $0.304_{0.032}$ \\
 & 16 & 10 & $0.909_{0.004}$ & $0.050_{0.008}$ & $0.325_{0.035}$ \\
 & 16 & 20 & $0.912_{0.004}$ & $0.042_{0.010}$ & $0.318_{0.014}$ \\
\bottomrule
\end{tabular}
}
\end{table}

\begin{table}
\small
\centering
\caption{
Accuracy and calibration of 
MAP vs. Laplace fine-tuning of LLaMA2-7B 
with posteriors computed at optimal MAP epochs.
}
\label{table:LLaMA_results_supp}
\resizebox{1.0\linewidth}{!}
{
\begin{tabular}{l|l|cc|cc|cc}
    \toprule
    \multirow{2}{*}{} & \multirow{2}{*}{Method} & \multicolumn{2}{c|}{OBQA @ epoch=3} & \multicolumn{2}{c|}{ARC-E @ epoch=5} & \multicolumn{2}{c}{ARC-C @ epoch=7} \\
    &     & MAP    & Bayesian    & MAP    & Bayesian    & MAP    & Bayesian \\
    \midrule
    \multirow{3}{*}{\rotatebox{90}{Acc $\uparrow$}} \rule{0pt}{2.25ex}

    
    & L-LoRA-XS* r=64    & $79.0_{1.1}$    & $79.4_{1.0}$    & $82.7_{0.2}$    & $83.4_{0.8}$    & $61.7_{4.3}$    & $62.7_{1.5}$ \\
    & L-LoRA-XS* r=32    & $77.0_{2.0}$    & $76.8_{2.0}$    & $82.9_{0.9}$    & $83.1_{0.7}$    & $53.6_{2.3}$    & $55.3_{1.7}$ \\
    \\
     
    \midrule
    \multirow{3}{*}{\rotatebox{90}{ECE $\downarrow$}} \rule{0pt}{2.25ex}

    
    & L-LoRA-XS* r=64    & $14.8_{1.0}$    & $5.9_{0.8}$    & $13.3_{0.6}$    & $6.3_{1.6}$    & $28.6_{3.5}$    & $7.5_{1.4}$ \\
    & L-LoRA-XS* r=32    & $15.3_{1.5}$    & $4.3_{0.4}$    & $12.2_{0.7}$    & $5.5_{1.0}$    & $29.2_{5.9}$    & $7.5_{2.7}$ \\
    \\
    
    \midrule
    \multirow{3}{*}{\rotatebox{90}{NLL $\downarrow$}} \rule{0pt}{2.25ex}

    
    & L-LoRA-XS* r=64    & $0.98_{0.07}$    & $0.64_{0.02}$    & $0.95_{0.03}$    & $0.54_{0.01}$    & $1.89_{0.27}$    & $0.99_{0.03}$ \\
    & L-LoRA-XS* r=32    & $0.90_{0.03}$    & $0.65_{0.02}$    & $0.81_{0.13}$    & $0.53_{0.02}$    & $1.58_{0.50}$    & $1.07_{0.03}$ \\
    \\
    
    \bottomrule
    \end{tabular}
}
\end{table}


\end{document}